\documentclass[runningheads]{llncs}
\usepackage{graphicx}

\usepackage{tikz}
\usepackage{comment}
\usepackage{amsmath,amssymb} 
\usepackage{color}
\usepackage[colorlinks,linkcolor=red, anchorcolor=blue, citecolor=green]{hyperref}
\usepackage[accsupp]{axessibility}  

\usepackage{longtable}

\usepackage{wrapfig}
\usepackage{multirow}
\usepackage{booktabs} 
\DeclareMathOperator*{\argmax}{arg\,max}

\newcommand{\eg}{\textit{e}.\textit{g}.}

\begin{document}
\pagestyle{headings}
\mainmatter
\def\ECCVSubNumber{2734}  

\title{Fine-Grained Scene Graph Generation with Data Transfer} 

\titlerunning{Fine-Grained Scene Graph Generation with Data Transfer}
%
\author{Ao Zhang\inst{1}$^*$ \and
Yuan Yao\inst{2}$^*$ \and
Qianyu Chen\inst{2} \and Wei Ji\inst{1}$^\dag$ 
\and Zhiyuan Liu\inst{2}$^\dag$, \\ 
 Maosong Sun\inst{2} \and Tat-Seng Chua\inst{1}}
\authorrunning{Zhang et al.}
%
\institute{Sea-NExT Joint Lab, Singapore\\
\hspace{0.5em}School of Computing, National University of Singapore, Singapore\\ \and
Department of Computer Science and Technology\\
Institute for Artificial Intelligence, Tsinghua University, Beijing, China\\
Beijing National Research Center for Information Science and Technology, China \\
\email{aozhang@u.nus.edu, yaoyuanthu@163.com}\\
}

{\let\thefootnote\relax\footnotetext{$*$ indicates equal contribution.}}
{\let\thefootnote\relax\footnotetext{$\dag$ Corresponding author: jiwei@nus.edu.sg, liuzy@tsinghua.edu.cn}}

\maketitle

\begin{abstract}
Scene graph generation (SGG) is designed to extract (\textit{subject}, \texttt{predicate}, \textit{object}) triplets in images. 
Recent works have made a steady progress on SGG, and provide useful tools for high-level vision and language understanding.
However, due to the data distribution problems including long-tail distribution and semantic ambiguity, the predictions of current SGG models tend to collapse to several frequent but uninformative predicates (\eg, \texttt{on}, \texttt{at}), which limits practical application of these models in downstream tasks. 
To deal with the problems above, we propose a novel Internal and External Data Transfer (IETrans) method, which can be applied in a plug-and-play fashion and expanded to large SGG with 1,807 predicate classes.
Our IETrans tries to relieve the data distribution problem by automatically creating an enhanced dataset that provides more sufficient and coherent annotations for all predicates. 
By applying our proposed method, a Neural Motif model doubles the macro performance for informative SGG.
The code and data are publicly available at \url{https://github.com/waxnkw/IETrans-SGG.pytorch}.
\keywords{Scene graph generation, Plug-and-play, Large-scale}
\end{abstract}

\section{Introduction}

\begin{figure}[t]
    \centering
    \includegraphics[width=0.7\columnwidth]{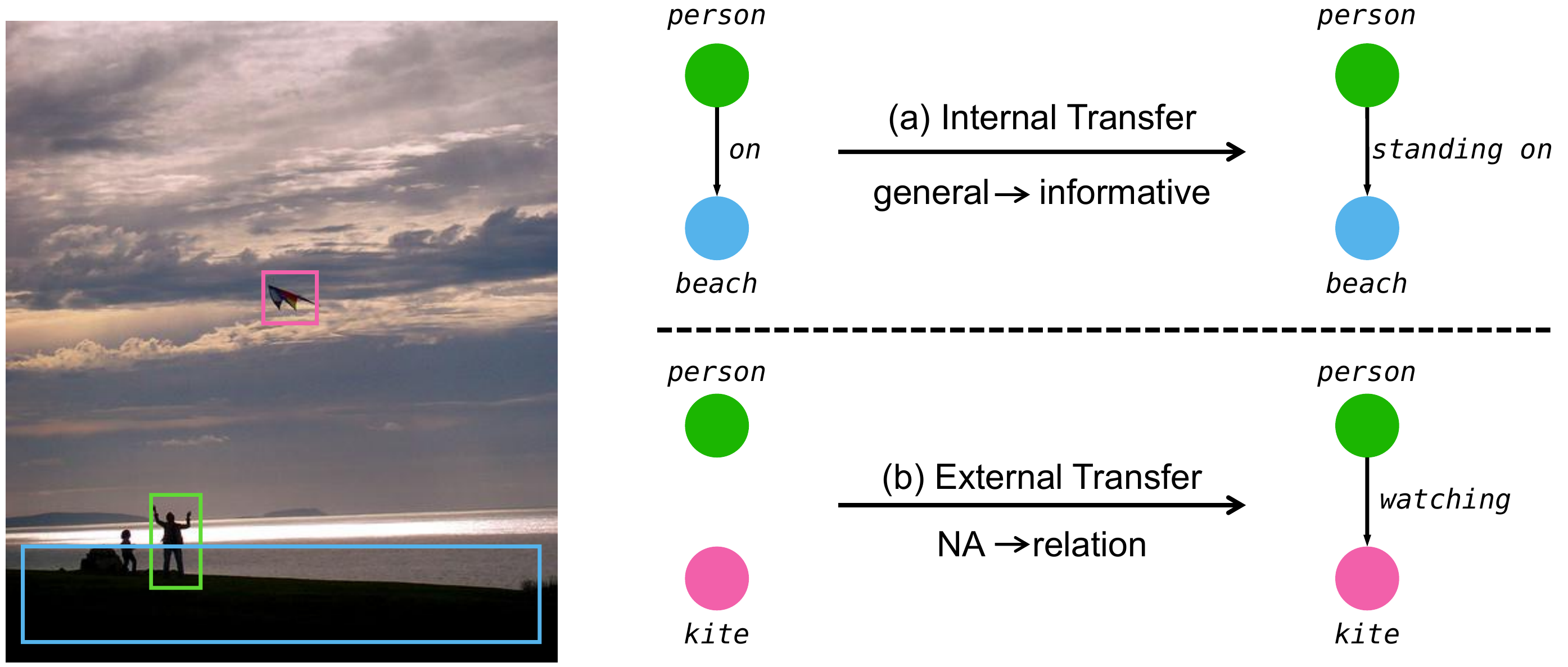}
    \caption{Generate an enhanced dataset automatically for better model training with: (a) \textbf{Internal Transfer}: Specify general predicate annotations as informative ones. (b) \textbf{External Transfer}: Relabel missed relations from \texttt{NA}.}
    \label{fig:example}
\end{figure}

Scene graph generation (SGG) aims to detect relational triplets (\eg, (\textit{man}, \texttt{riding}, \textit{bike})) in images. 
As an essential task for connecting vision and language, it can serve as a fundamental tool for high-level vision and language tasks, such as visual question answering~\cite{vqa2015antol,sggforvqa2017teney,xiao2022video,li2022invariant}, image captioning~\cite{auto2019yang,unpaired2019gu}, and image retrieval~\cite{imrt2015johnson,cross2020wang,wei2022rethinking}. However, existing SGG methods can only make correct predictions on a limited number of predicate classes (\eg, 29 out of 50 pre-defined classes~\cite{zellers2018motif}), among which a majority of predicates are trivial and uninformative (\eg, \texttt{on}, \texttt{and}, \texttt{near}).
This undermines the application of SGG for downstream tasks.
To address the limitation, we first identify two main problems that need to deal with:

\begin{itemize}
    \item[$\bullet$] \textbf{Long-tail problem:} the problem refers to the phenomenon that annotations mainly concentrate on a few head predicate classes, and are much sparse in most tail predicate classes. For example, in Visual Genome\cite{krishna2017visual}, there are over 100K samples for the top 5 predicate classes, while over 90\% of predicate classes have less than 10 samples. As a result, the performance of tail predicate classes is poor due to the lack of effective supervision.
    \item[$\bullet$] \textbf{Semantic ambiguity:} many samples can be described as either general predicate class (\eg, \texttt{on}) or an informative one (\eg, \texttt{riding}). However, data annotators prefer some general (and thus uninformative) predicate classes to informative ones for simplicity. This causes conflicts in the widely adopted single-label optimization since different labels are annotated for the same type of instances. Thus, even when the informative predicates have enough training samples, the prediction will easily collapse to the general ones.
\end{itemize}

To address the problems mentioned above, recent works propose to use resampling~\cite{desai2021learning,li2021bipartite}, reweighting~\cite{yan2020pcpl}, and post-processing methods~\cite{tang2020unbiased,guo2021general}. However, we argue that these problems can be better alleviated by enhancing the existing dataset into a reasonable dataset, that contains more abundant training samples for tail classes and also provides coherent annotations for different classes.

To this end, we propose a novel framework named \textbf{I}nternal and \textbf{E}xternal data \textbf{Trans}fer (\textbf{IETrans}), which can be equipped to different baseline models in a plug-and-play applied in a fashion.
As shown in Figure~\ref{fig:example}, we automatically transfer data from general predicates to informative ones (\textbf{Internal Transfer}) and relabel relational triplets missed by annotators (\textbf{External Transfer}).
(1) \textbf{For internal transfer}, we first identify the general-informative relational pairs based on the confusion matrix, and then conduct a triplet-level data transfer from general ones to informative ones.
The internal transfer will not only alleviate the optimization conflict caused by semantic ambiguity but also provide more data for tail classes;
(2) \textbf{For external transfer}, there exist many positive samples missed by annotators~\cite{krishna2017visual,lu2016visual}, which are usually treated as negative samples by current methods.
However, this kind of data can be considered as a potential data source, covering a wide range of predicate categories.
Inspired by Visual Distant Supervision~\cite{yao2021visual} which employs \texttt{NA} samples for pre-training, we also consider the \texttt{NA} samples, which are the union of negative and missed annotated samples.
The missed annotated samples can be relabeled to provide more training samples.

It is worth noting that both internal transfer and external transfer are indispensable for improving SGG performance.
Without the internal transfer, the external transfer will suffer from the semantic ambiguity problem.
Meanwhile, the external transfer can further provide training samples for tail classes, especially for those that have weak semantic connection with head classes.

Exhaustive experiments show that our method is both adaptive to different baseline models and expansible to large-scale SGG.
We equip our data augmentation method with 4 different baseline models and find that it can significantly boost all models' macro performance and achieve SOTA performance for F@K metric, a metric for overall evaluation.
For example, a Neural Motif Model with our proposed method can double the mR@100 performance and achieve the highest F@100 among all model-agnostic methods on predicate classification task of the widely adopted VG-50~\cite{xu2017scene} benchmark.

To validate the scalability of our proposed method, we additionally propose a new benchmark with 1,807 predicate classes (VG-1800), which is more practical and challenging. 
To provide a reliable and stable evaluation, we manually remove unreasonable predicate classes and make sure there are over 5 samples for each predicate class on the test set.
On VG-1800, our method achieves SOTA performance with significant superiority compared with all baselines.
The proposed IETrans can make correct predictions on 467 categories, compared with all other baselines that can only correctly predict less than 70 categories.
While the baseline model can only predict relations like (\textit{cloud}, \texttt{in}, \textit{sky}) and (\textit{window}, \texttt{on}, \textit{building}), our method enables to generate informative ones like (\textit{cloud}, \texttt{floating through}, \textit{sky}) and (\textit{window}, \texttt{on exterior of}, \textit{building}).

Our main contributions are summarized as follows: (1) To cope with the long-tail problem and semantic ambiguity in SGG, we propose a novel IETrans framework to generate an enhanced training set, which can be applied in a plug-and-play fashion.
(2) We propose a new VG-1800 benchmark, which can provide reliable and stable evaluation for large-scale SGG.
(3) Comprehensive experiments demonstrate the effectiveness of our IETrans in training SGG models.

\section{Related Works}
\subsection{Scene Graph Generation}
As an important tool of connecting vision and language, SGG~\cite{xu2017scene,lu2016visual,li2021interventional} has drawn widespread attention from the community. SGG is first proposed as visual relation detection (VRD)~\cite{lu2016visual}, in which each relation is detected independently. Considering that relations are highly dependent on their context, \cite{xu2017scene} further proposes to formulate VRD as a dual-graph generation task, which can incorporate context information.
Based on \cite{xu2017scene}, different methods~\cite{lin2020gps,tang2019learning,zellers2018motif} are proposed to refine the object and relation representations in the scene graph.
For example, \cite{lin2020gps} proposes a novel message passing mechanism that can encode edge directions into node representations. 
Recently, CPT~\cite{yao2021cpt} and PEVL~\cite{yao2022pevl} propose to employ pre-trained vision-language models for SGG.
CPT shows promising few-shot ability and PEVL shows much better performance than models training from scratch.

\subsection{Informative Scene Graph Generation}
Although making steady progress on improving recall on SGG task, \cite{tang2019learning,chen2019knowledge} point out that the predictions of current SGG models are easy to collapse to several general and trivial predicate classes.
Instead of only focusing on recall metric, \cite{tang2019learning,chen2019knowledge} propose a new metric named mean recall, which is the average recall of all predicate classes.
\cite{tang2020unbiased} employs a causal inference framework, which can eliminate data bias during the inference process.
CogTree~\cite{yu2020cogtree} proposes to leverage the semantic relationship between different predicate classes, and design a novel CogTree loss to train models that can make informative predictions.
In BGNN~\cite{li2021bipartite}, the authors design a bi-level resampling strategy, which can help to provide a more balanced data distribution during the training process. 
However, previous works of designing new loss or conducting resampling, only focus on predicate-level adjustment, while the visual relation is triplet-level.
For example, given the subject \textit{man} and object \textit{skateboard}, the predicate \texttt{riding} is an informative version of \texttt{standing on}, while given the subject \textit{man} and object \textit{horse}, \texttt{riding} will not be an informative alternative of \texttt{standing on}.
Thus, instead of using less precise predicate-level manipulation, we employ a triplet-level transfer.

\subsection{Large-scale Scene Graph Generation}
In the last few years, there are some works~\cite{abdelkarim2020long,zhuang2018hcvrd,zhang2019large,yao2021visual} focusing on large-scale SGG. 
Then, how to provide a reliable evaluation is an important problem.
\cite{zhang2019large} first proposes to study large-scale scene graph generation and makes a new split of Visual Genome dataset named VG80K, which contains 29,086 predicate categories.
However, the annotations are too noisy to provide reliable evaluation.
To cope with this problem, \cite{abdelkarim2020long} further cleans the dataset, and finally reserves 2,000 predicate classes.
However, only 1,429 predicate classes are contained in the test set, among which 903 relation categories have no more than 5 samples.
To provide enough samples for each predicate class' evaluation, we re-split the Visual Genome to ensure each predicate class on the test set has more than 5 samples, and the total predicate class number is 1,807.
For the proposed methods, \cite{zhang2019large} employs a triplet loss to regularize the visual representation with the constraint on word embedding space.
RelMix~\cite{abdelkarim2020long} proposes to conduct data augmentation with the format of feature mixup.
Visual distant supervision~\cite{yao2021visual} pre-trains the model on relabeled \texttt{NA} data with the help of a knowledge base and achieve significant improvement on a well-defined VG setting without semantic ambiguity. 
However, the data extension will be significantly limited by the semantic ambiguity problem. 
To deal with this problem, we propose an internal transfer method to generate informative triplets.

\section{Method}
\begin{figure*}[t]
    \centering
    \includegraphics[width=\textwidth]{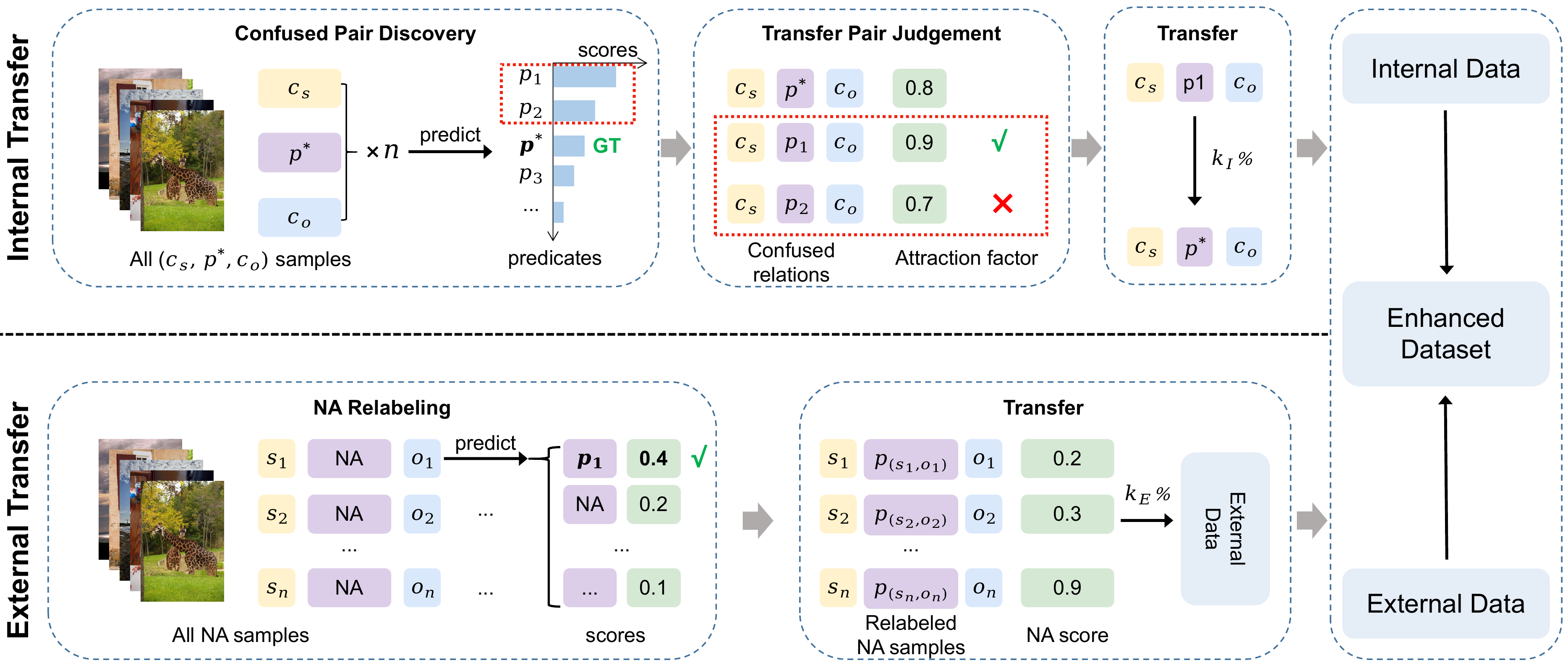}
   \caption{Illustration of our proposed IETrans to generate an enhanced dataset. \textbf{Internal transfer} is designed to transfer data from general predicate to informative ones. \textbf{External transfer} is designed to relabel \texttt{NA} data. To avoid misunderstanding, $(c_s, p^*, c_o)$ is a relational triplet class.
   $(s_i, p_{(s_i, o_i)}, o_i)$ represents a single relational triplet instance.}
    \label{fig:method}
\end{figure*}

In this section, we first introduce the internal data transfer and external data transfer, respectively, and then elaborate how to utilize them collaboratively. Figure~\ref{fig:method} shows the pipeline of our proposed IETrans.

The goal of our method is to generate an enhanced dataset automatically, which should provide more training samples for tail classes, and also specify general predicate classes as informative ones.
Concretely, as shown in Figure~\ref{fig:example}, the general relation \texttt{on} between (\textit{person}, \textit{beach}) need to be specified as more informative one \texttt{standing on}, and the missed annotations between (\textit{person}, \textit{kite}) can be labeled so as to provide more training samples.

\subsection{Problem Definition}

\smallskip
\noindent
\textbf{Scene Graph Generation.} Given an image $I$, a scene graph corresponding to $I$ has a set of objects $O=\{(b_i, c_i)\}_{i=1}^{N_o}$ and a set of relational triplets $E=\{(s_i, p_{(s_i, o_i)}, o_i)\}_{i=1}^{N_e}$. For each object $(b_i, c_i)$, it consists of an object bounding box $b_i \in \mathbb{R}^4$ and an object class $c_i$ which belongs to the pre-defined object class set $\mathcal{C}$. With $s_i \in O$ and $o_i \in O$, $p_i$ is defined as relation between them and belongs to the pre-defined predicate class set $\mathcal{P}$.

\smallskip
\noindent
\textbf{Inference.} SGG is defined as a joint detection of objects and relations.
Generally, an SGG model will first detect the objects in the image $I$.
Based on the detected objects, a typical SGG model will conduct a feature refinement for objects and relation representation, and then classify the objects and relations.

\subsection{Internal Data Transfer}
The key insight of internal transfer is to transfer samples from general predicate classes to their corresponding informative ones, like the example shown in Figure~\ref{fig:example}.
We split the process into 3 sub-steps, including \textbf{(1) Confusion Pair Discovery}: specify confused predicate pairs as potential general-informative pairs for given subject and object classes. 
\textbf{(2) Transfer Pair Judgement}: judge whether the candidate pair is valid. \textbf{(3) Triplet Transfer}: transfer data from the selected general predicate class to the corresponding informative one.

\smallskip
\noindent
\textbf{Confusion Pair Discovery.}
To find general predicate classes and corresponding informative ones, a straightforward way is to annotate the possible relation transitions manually.
However, relations are highly dependent on the subject and object classes, i.e. relation is triplet-level rather than predicate-level.
For example, given the entity pair \textit{man} and \textit{bike}, \texttt{riding} is a sub-type of \texttt{sitting on}, while for \textit{man} and \textit{skateboard}, \texttt{riding} shares different meaning with \texttt{sitting on}.
In this condition, even under 50 predicate classes settings, the possible relation elements will scale up to an infeasible number for human annotation.
Another promising alternative is to employ pre-defined knowledge bases, such as WordNet~\cite{miller1995wordnet} and VerbNet~\cite{kipper2006extending}.
However, existing knowledge bases are not specifically designed to cope with visual relation problems, which result in a gap between visual and textual hierarchies~\cite{wan2020nbdt}.

\begin{wrapfigure}{R}{0.4\textwidth}
        \centering
        \includegraphics[width=1\linewidth]{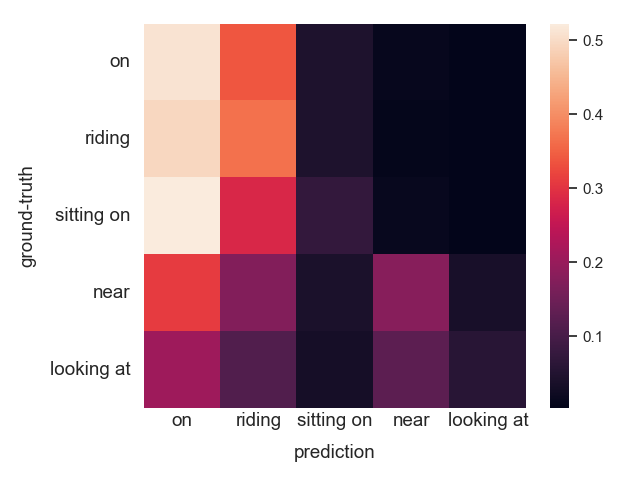}
        \caption{Confusion matrix of Motif~\cite{zellers2018motif}'s prediction score on all entity pairs in VG training set with the subject \textit{man} and the object \textit{motorcycle}.}
        \label{fig:confusion}
\end{wrapfigure}

Thus, in this work, we try to specify the general-informative pairs by taking advantage of information within the dataset, and leave the exploration of external knowledge sources for future work.
A basic observation is that \textbf{informative predicate classes are easily confused by general ones}.
Thus, we can first find confusion pairs as candidate general-informative pairs.
By observing the predictions of the pre-trained  Motif~\cite{zellers2018motif} model, we find that the collapse from informative predicate classes to general ones, appears not only on the test set but also on the training set.
As shown in Figure~\ref{fig:confusion}, the predicate classes \texttt{riding} and \texttt{sitting on} are significantly confused by a more general predicate class \texttt{on}.

On the training set, given a relational triplet class $(c_s, p, c_o)$, we use a pre-trained baseline model to predict predicate labels of all samples belonging to $(c_s, p, c_o)$, and average their score vectors.
We denote the aggregated scores for all predicates as $S = \{s_{p_i} | p_i \in \mathcal{P} \}$.
From $S$, we select all predicate classes with higher prediction scores than the ground-truth annotation $p$, which can be formulated as $\mathcal{P}_c = \{p_i | s_{p_i}>s_{p}\}$.
$\mathcal{P}_c$ can be considered as the most confusing predicate set for $(c_s, p, c_o)$, which can serve as candidate transfer sources.

\smallskip
\noindent
\textbf{Transfer Pair Judgement.}
However, a confused predicate class does not equal to a general one.
Sometimes, a general predicate can also be confused by an informative predicate.
For example, in Figure~\ref{fig:confusion}, under the constraint of $c_s=man$ and $c_o=motorcycle$, the less informative predicate \texttt{sitting on} is confused by the more informative predicate \texttt{riding}.
In this condition, it is not a good choice to transfer from \texttt{riding} to \texttt{sitting on}.
Thus, we need to further select the truly general predicates from the candidate set $\mathcal{P}_c$.

To select the most possible general predicate classes from $\mathcal{P}_c$, we first introduce an important feature that is useful to recognize general predicate classes. As observed by \cite{yu2020cogtree}, \textbf{the general predicate classes usually cover more diverse relational triplets}, while informative ones are limited.
Based on this observation, we can define the attraction factor of a triplet category $(c_s, p, c_o)$ as:
\begin{equation}
    A(c_s, p, c_o) = \frac{1}{\sum_{c_i, c_j \in \mathcal{C}}{\mathcal{I}(c_i, p, c_j)}},
\end{equation}
where $\mathcal{C}$ is the object categories set and $\mathcal{I}(t)$ indicates whether the triplet category $t$ exists in the training set, which can be formulated as:
\begin{equation}
    \mathcal{I}(t) = 
    \begin{cases}
    1, &\text{if $t \in$ training set}\\
    0, &\text{otherwise}
    \end{cases}
\end{equation}
The denominator of $A(c_s, p_i, c_o)$ is the number of relational triplet types containing $p_i$. 
Thus, $A(c_s, p_i, c_o)$ with smaller value means $p_i$ is more likely to be a general predicate.
Concretely, when $A(c_s, p_i, c_o) < A(c_s, p, c_o)$, we transfer data from $(c_s, p_i, c_o)$ to $(c_s, p, c_o)$.

However, only considering the number of relational triplet types also has drawbacks: 
some relational triplets with very limited number of samples (\eg, only 1 or 2 samples) might be annotation noise, while these relational triplets are easily selected as transfer targets.
Transferring too much data to such uncertain relational triplets will significantly degenerate models' performance.
Thus, we further consider the number of each relational triplet and modify the attraction factor as:
\begin{equation}
    A(c_s, p, c_o) = \frac{N(c_s, p, c_o)}{\sum_{c_i, c_j \in \mathcal{C}}{\mathcal{I}(c_i, p, c_j) \cdot N(c_i, p, c_j) }},
\end{equation}
where $N(t)$ denotes the number of instances with relational type $t$ in the training set.
With the attraction factor, we can further filter the candidate confusion set $\mathcal{P}_c$ to the valid transfer source for $(c_s, p, c_o)$: 
\begin{equation}
    \mathcal{P}_s=\{p_i| (p_i \in \mathcal{P}_c) \wedge (A(c_s, p_i, c_o) < A(c_s, p, c_o)) \},
\end{equation}
where $\wedge$ denotes the logical conjunction operator.

\smallskip
\noindent
\textbf{Triplet Transfer.}
Given the transfer source $\mathcal{P}_s$, we collect all samples in the training set satisfying:
\begin{equation}
    T= \{(o_i, p_k, o_j) | (c_{o_i}=c_s) \wedge (p_k \in \mathcal{P}_s) \wedge (c_{o_j}=c_o) \}.
\end{equation}
Then, we sort $T$ by model's prediction score of $p$, and transfer the top $k_I$\% samples to the target triplet category $(c_s, p, c_o)$.
Note that, a triplet instance may need to be transferred to more than one relational triplets.
To deal with the conflict, we choose the target predicate with the highest attraction factor.

\subsection{External Data Transfer}
The goal of our external transfer is to relabel unannotated samples to excavate missed relational triplets, as the example shown in Figure~\ref{fig:example}.

\smallskip
\noindent
\textbf{NA Relabeling.}
\texttt{NA} samples refer to all unannotated samples in the training set, including both truly negative samples and missed positive samples.
In external transfer, \texttt{NA} samples are directly considered as the transfer source and are relabeled as existing relational triplet types.

To get the \texttt{NA} samples, we first traverse all unannotated object pairs in images. 
However, considering that data transfer from all \texttt{NA} samples to all possible predicate classes will bring heavy computational burden, and inevitably increase the difficulty of conducting precise transfer, so as to sacrifice the quality of  transferred data.
Thus, we only focus on object pairs whose bounding boxes have overlaps and limit the possible transfer targets to existing relational triplet types.
The exploration of borrowing zero-shot relational triplets from \texttt{NA} is left for future work.

Given a sample $(s, \texttt{NA}, o)$, we can get its candidate target predicate set as:
\begin{equation}
    \text{Tar}(s, \texttt{NA}, o) = \{p |(p \in \mathcal{P}) \wedge  ({N(c_s, p, c_o)} > 0) \wedge (\text{IoU}(b_{s}, b_{o})>0) \},
\end{equation}
where $\mathcal{P}$ denotes pre-defined predicate classes, $b_s$ and $b_o$ denote bounding boxes of $s$ and $o$, and $\text{IoU}$ denotes the intersection over union.

Given a triplet $(s, \texttt{NA}, o)$, the predicate class with the highest prediction score except for \texttt{NA} is chosen.
The label assignment can be formulated as:
\begin{equation}
    p_{(s, o)} = \argmax_{p\in \text{Tar}(s, \texttt{NA}, o)}(\phi^{p}(s, o)),
\end{equation}
where $\phi^{p}(\cdot)$ denotes the prediction score of predicate $p$.

\smallskip
\noindent
\textbf{NA Triplet Transfer.}
To decide transfer or not, we rank all chosen $(s, \texttt{NA}, o)$ samples according to \texttt{NA} scores in an ascending order.
The lower \texttt{NA} score means the sample is more likely to be a missed positive sample. 
Similar with internal transfer, we simply transfer the top $k_E$\% data.

\subsection{Integration}
Internal transfer is conducted on annotated data and external transfer is conducted on unannotated data, which are orthogonal to each other.
Thus, we can simply merge the data without conflicts.
After obtaining the enhanced dataset, we re-train a new model from scratch and use the new model to make inferences on the test set.

\section{Experiments}
In this section, we first show the generalizability of our method with different baseline models and the expansibility to large-scale SGG. 
We also make ablation studies to explore the influence of different modules and hyperparameters.
Finally, analysis is conducted to show the effectiveness of our method in enhancing the current dataset.

\subsection{Generalizability with Different Baseline Models}
We first validate the generalizability of our method with different baseline models and its effectiveness when compared with current SOTA methods.

\smallskip
\noindent
\textbf{Datasets.} Popular VG-50~\cite{xu2017scene} benchmark is employed, which consists of 50 predicate classes and 150 object classes.

\smallskip
\noindent
\textbf{Tasks.}
Following previous works~\cite{xu2017scene,tang2020unbiased,zellers2018motif}, we evaluate our model on three widely used SGG tasks:
(1) \textbf{Predicate Classification (PREDCLS)} provides both localization and object classes, and requires models to recognize predicate classes.
(2) \textbf{Scene Graph Classification (SGCLS)} provides only correct localization and asks models to recognize both object and predicate classes.
(3) In \textbf{Scene Graph Detection (SGDET)}, models are required to first detect the bounding boxes and then recognize both object and predicate classes.

\smallskip
\noindent
\textbf{Metrics.}
Following previous works~\cite{yu2020cogtree,tang2019learning}, we use Recall@K (\textbf{R@K}) and mean Recall@K (\textbf{mR@K}) as our metrics.
However, different trade-offs between R@K and mR@K are made in
different methods, which makes it hard to make a direct comparison.
Therefore, we further propose an overall metric \textbf{F@K} to jointly evaluate R@K and
mR@K, which is the harmonic average of R@K and mR@K.

\smallskip
\noindent
\textbf{Baselines.}
We categorize several baseline methods into two categories:
(1) \textbf{Model-agnostic baselines.} They refers to methods that can be applied in a plug-and-play fashion. For this part, we include Resampling~\cite{li2021bipartite},  TDE~\cite{tang2020unbiased}, CogTree~\cite{yu2020cogtree}, EBM~\cite{ebm2021energy}, DeC~\cite{dec2021semantic}, and DLFE~\cite{dlfe2021recovering}.
(2) \textbf{Specific models.} We also include some dedicated designed models with strong performance, including KERN~\cite{kern2019knowledge}, KERN~\cite{kern2019knowledge}, GBNet~\cite{gbnet2020bridging}, BGNN~\cite{li2021bipartite}, DT2-ACBS~\cite{desai2021learning}, and PCPL~\cite{yan2020pcpl}.

\smallskip
\noindent
\textbf{Implementation Details.}
Following \cite{tang2020unbiased}, we employ a pre-trained Faster-RCNN~\cite{ren2015faster} with ResNeXt-101-FPN~\cite{lin2017feature,xie2017aggregated} backbone.
In the training process, the parameters of the detector are fixed to reduce the computation cost.
The batch size is set to 12, and the learning rate is 0.12, except for Transformer.
We optimize all models with an SGD optimizer. Specifically, to better balance the data distribution, the external transfer will not be conducted for the top 15 frequent predicate classes.
To avoid deviating too much from the original data distribution, the frequency bias item calculated from the original dataset is applied to our IETrans in the inference stage.
For internal and external transfer, the $k_I$ is set to 70\% and $k_E$ is set to 100\%.
Please refer to the Appendix for more details.

\setlength{\tabcolsep}{3pt}
\begin{table*}[t]
    \begin{center}
    \caption{Performance (\%) of our method and other baselines on VG-50 dataset. \textbf{IETrans} denotes different models equipped with our IETrans. \textbf{Rwt} denotes using the reweighting strategy.\label{label:vg50}}
    \resizebox{\linewidth}{!}{%
    \small
    \begin{tabular}{ll ccc ccc ccc}
    \toprule
    &  \multirow{2}{*}{Models} & \multicolumn{3}{c}{Predicate Classification}& \multicolumn{3}{c}{Scene Graph Classification} & \multicolumn{3}{c}{Scene Graph Detection} \\
    \cmidrule(lr){3-5} \cmidrule(lr){6-8} \cmidrule(lr){9-11}
    &  & R@50 / 100 & mR@50 / 100 & F@50 / 100 & R@50 / 100 & mR@50 / 100 & F@50 / 100 & R@50 / 100 & mR@50 / 100 & F@50 / 100 \\
    \midrule
    
 \parbox[t]{5mm}{\multirow{5}{*}{\rotatebox[origin=c]{90}{Specific}}}
  &  KERN~\cite{kern2019knowledge}  & 65.8 / 67.6 & 17.7 / 19.2 & 27.9 / 29.9 & 36.7 / 37.4 & 9.4 / 10.0 & 15.0 / 15.8 & 27.1 / 29.8 & 6.4 / 7.3 & 10.4 / 11.7  \\
  &  GBNet~\cite{gbnet2020bridging} & 66.6 / 68.2 & 22.1 / 24.0 & 33.2 / 35.5 & 37.3 / 38.0 & 12.7 / 13.4 & 18.9 / 19.8 & 26.3 / 29.9 & 7.1 / 8.5 & 11.2 / 13.2  \\
 &  BGNN~\cite{li2021bipartite} & 59.2 / 61.3 & 30.4 / 32.9 & 40.2 / 42.8 & 37.4 / 38.5 & 14.3 / 16.5 & 20.7 / 23.1 & 31.0 / 35.8 & 10.7 / 12.6 & 15.9 / 18.6
  \\
 &  DT2-ACBS~\cite{desai2021learning} & 23.3 / 25.6 & 35.9 / 39.7 & 28.3 / 31.1 & 16.2 / 17.6 & 24.8 / 27.5 & 19.6 / 21.5 & 15.0 / 16.3 & 22.0 / 24.0 & 17.8 / 19.4  \\
 &  PCPL~\cite{yan2020pcpl} & 50.8 / 52.6 & 35.2 / 37.8 & 41.6 / 44.0 & 27.6 / 28.4 & 18.6 / 19.6 & 22.2 / 23.2 & 14.6 / 18.6 & 9.5 / 11.7 & 11.5 / 14.4 \\
    \midrule
 \parbox[t]{5mm}{\multirow{23}{*}{\rotatebox[origin=c]{90}{Model-Agnostic}}}
 &  Motif~\cite{zellers2018motif} & 64.0 / 66.0 & 15.2 / 16.2 & 24.6 / 26.0 & 38.0 / 38.9 & 8.7 / 9.3 & 14.2 / 15.0 & 31.0 / 35.1 & 6.7 / 7.7 & 11.0 / 12.6  \\
  &  \quad -TDE~\cite{tang2020unbiased} & 46.2 / 51.4 & 25.5 / 29.1 & 32.9 / 37.2 & 27.7 / 29.9 & 13.1 / 14.9  & 17.8 / 19.9 & 16.9 / 20.3 & 8.2 / 9.8 & 11.0 / 13.2  \\
 & \quad -CogTree~\cite{yu2020cogtree} & 35.6 / 36.8 & 26.4 / 29.0 & 30.3 / 32.4 & 21.6 / 22.2 & 14.9 / 16.1 & 17.6 / 18.7 & 20.0 / 22.1 & 10.4 / 11.8 & 13.7 / 15.4  \\
 &  \quad -EBM~\cite{ebm2021energy} & - / - & 18.0 / 19.5 & - / - & - / - & 10.2 / 11.0 & - / - & - / - & 7.7 / 9.3 & - / -  \\
   & \quad -DeC~\cite{dec2021semantic} & - / - & 35.7 / 38.9 & - / - & - / - & 18.4 / 19.1 & - / - & - / - & 13.2 / 15.6 & - / -  \\
  &  \quad -DLFE~\cite{dlfe2021recovering} & 52.5 / 54.2 & 26.9 / 28.8 & 35.6 / 37.6 & 32.3 / 33.1 & 15.2  / 15.9 & 20.7 / 21.5 & 25.4 / 29.4 & 11.7  / 13.8 & 16.0 / 18.8  \\
  & \quad -\textbf{IETrans (ours)} & 54.7 / 56.7 & 30.9 / 33.6 & 39.5 / 42.2 & 32.5 / 33.4 & 16.8 / 17.9 & 22.2 / 23.3 & 26.4 / 30.6 & 12.4 / 14.9 & 16.9 / 20.0   \\
 &  \quad -\textbf{IETrans+Rwt (ours)} & 48.6 / 50.5 & \textbf{35.8} / \textbf{39.1} & \textbf{41.2} / \textbf{44.1} & 29.4 / 30.2 & \textbf{21.5} / \textbf{22.8} & \textbf{24.8} / \textbf{26.0} & 23.5 / 27.2 & \textbf{15.5} / \textbf{18.0} & \textbf{18.7} / \textbf{21.7}  \\
 \cmidrule{2-11}
&  VCTree~\cite{tang2019learning} & 64.5 / 66.5 & 16.3 / 17.7 & 26.0 / 28.0 & 39.3 / 40.2 & 8.9 / 9.5  & 14.5 / 15.4 & 30.2 / 34.6 &  6.7 / 8.0 & 11.0 / 13.0   \\ 
 &  \quad -TDE~\cite{tang2020unbiased} & 47.2 / 51.6 & 25.4 / 28.7 & 33.0 / 36.9 & 25.4 / 27.9 & 12.2 / 14.0 & 16.5 / 18.6 & 19.4 / 23.2 & 9.3 / 11.1 & 12.6 / 15.0   \\
 &  \quad -CogTree~\cite{yu2020cogtree} & 44.0 / 45.4  & 27.6 / 29.7 & 33.9 / 35.9 & 30.9 / 31.7 & 18.8 / 19.9 & 23.4 / 24.5 & 18.2 / 20.4 & 10.4 / 12.1 & 13.2 / 15.2  \\
  &  \quad -EBM~\cite{ebm2021energy} & - / - & 18.2 / 19.7 & - / - & - / - & 12.5 / 13.5 & - / - & - / - & 7.7 / 9.1 & - / -  \\
 &  \quad -DLFE~\cite{dlfe2021recovering} & 51.8 / 53.5 & 25.3 / 27.1 & 34.0 / 36.0 & 33.5 / 34.6 & 18.9  / 20.0 & \textbf{24.2} / 25.3 & 22.7 / 26.3 & 11.8  / 13.8 & 15.5 / 18.1   \\
  &  \quad -\textbf{IETrans (ours)} & 53.0 / 55.0 & 30.3 / 33.9 & 38.6 / 41.9 & 32.9 / 33.8 & 16.5 / 18.1 & 22.0 / 23.6 & 25.4 / 29.3 & 11.5 / 14.0 & 15.8 / 18.9  \\
 & \quad -\textbf{IETrans+Rwt (ours)} & 48.0 / 49.9 & \textbf{37.0} / \textbf{39.7} & \textbf{41.8} / \textbf{44.2} & 30.0 / 30.9 & \textbf{19.9} / \textbf{21.8} & 23.9 / \textbf{25.6} & 23.6 / 27.8 & \textbf{12.0} / \textbf{14.9} & \textbf{15.9} / \textbf{19.4}  \\
\cmidrule{2-11}
& GPS-Net~\cite{lin2020gps} & 65.1 / 66.9 & 15.0 / 16.0 & 24.4 / 25.8 & 36.9 / 38.0 & 8.2 / 8.7 & 13.4 / 14.2 &  30.3 / 35.0  & 5.9 / 7.1 & 9.9 / 11.8   \\
  & \quad -Resampling~\cite{li2021bipartite} & 64.4 / 66.7 & 19.2 / 21.4 & 29.6 / 32.4 & 37.5 / 38.6 & 11.7 / 12.5 & 17.8 / 18.9 & 27.8 / 32.1 & 7.4 / 9.5 & 11.7 / 14.7   \\
  & \quad -DeC~\cite{dec2021semantic} & - / - & \textbf{35.9} / 38.4 & - / - & - / - & 17.4 / 18.5 & - / - & - / - & 11.2 / 15.2 & - / -  \\

 &  \quad -\textbf{IETrans (ours)} & 52.3 / 54.3 & 31.0 / 34.5 & 38.9 / 42.2 & 31.8 / 32.7 & 17.0 / 18.3 & 22.2 / 23.5 & 25.9 / 28.1 & 14.6 / 16.5 & 18.7 / 20.8  \\
  &  \quad -\textbf{IETrans+Rwt (ours)} & 47.5 / 49.4 & 34.9 / \textbf{38.6} & \textbf{40.2} / \textbf{43.3} & 29.3 / 30.3 & \textbf{19.8} / \textbf{21.6} & \textbf{23.6} / \textbf{25.2} & 23.1 / 25.0 & \textbf{16.2} / \textbf{18.8} & \textbf{19.0} / \textbf{21.5} \\
\cmidrule{2-11}
&  Transformer~\cite{tang2020unbiased} & 63.6 / 65.7 & 17.9 / 19.6 & 27.9 / 30.2 & 38.1 / 39.2 & 9.9 / 10.5 & 15.7 / 16.6 & 30.0 / 34.3 & 7.4 / 8.8 & 11.9 / 14.0  \\
 &  \quad -CogTree~\cite{yu2020cogtree} & 38.4 / 39.7  & 28.4 / 31.0  & 32.7 / 34.8 & 22.9 / 23.4 & 15.7 / 16.7 & 18.6 / 19.5 & 19.5 / 21.7 & 11.1 / 12.7  & 14.1 / 16.0 \\
&  \quad -\textbf{IETrans (ours)} & 51.8 / 53.8 & 30.8 / 34.7 & 38.6 / 42.2 & 32.6 / 33.5 & 17.4 / 19.1 & 22.7 / 24.3 & 25.5 / 29.6 & 12.5 / 15.0 & 16.8 / 19.9  \\
 &  \quad -\textbf{IETrans+Rwt (ours)} & 49.0 / 50.8 & \textbf{35.0} / \textbf{38.0} & \textbf{40.8} / \textbf{43.5} & 29.6 / 30.5 & \textbf{20.8} / \textbf{22.3} & \textbf{24.4} / \textbf{25.8} & 23.1 / 27.1 & \textbf{15.0} / \textbf{18.1} & \textbf{18.2} / \textbf{21.7} \\
    \bottomrule
    \end{tabular}
    }
    \end{center}
\end{table*}

\smallskip
\noindent
\textbf{Comparison with SOTAs.}
We report the results of our IETrans and baselines for VG-50 in Table ~\ref{label:vg50}.
Based on the observation of experimental results, we have summarized the following conclusions:

\textbf{Our IETrans is adaptive to different baseline models.}
We equip our method with 4 different models, including Motif~\cite{zellers2018motif}, VCTree~\cite{tang2019learning}, GPS-Net~\cite{lin2020gps}, and Transformer~\cite{tang2020unbiased}.
The module architectures range from conventional CNN to TreeLSTM (VCTree) and self-attention layers (Transformer).
The training algorithm contains both supervised training and reinforcement learning (VCTree).
Despite the model diversity, our IETrans can boost all models' mR@K metric and also achieve competitive F@K performance.
For example, our IETrans can double mR@50/100 and improve the overall metric F@50/100 for over 9 points across all 3 tasks for GPS-Net.

\textbf{Compared with other model-agnostic methods, our method outperforms all of them in nearly all metrics.}
For example, when applying IETrans to Motif on PREDCLS, our model can achieve the highest R@50/100 and mR@50/100 among all model-agnostic baselines except for DeC.
After adding the reweighting strategy, our IETrans can outperform DeC on mR@K.

\textbf{Compared with strong specific baselines, our method can also achieve competitive performance on mR@50/100, and best overall performance on F@50/100.}
Considering mR@50/100, our method with reweighting strategy is slightly lower than DT2-ACBS on SGCLS and SGDET tasks, while our method performs much better than them on R@50/100 (\eg, 24.3 points of VCTree on PREDCLS task).
For overall comparison considering F@50/100 metrics, our VCTree+IETrans+Rwt can achieve the best F@50/100 on PREDCLS and Motif+IETrans+Rwt achieves the best F@50/100 in SGCLS and SGDET task.

\subsection{Expansibility to Large-Scale SGG}
We also validate our IETrans on VG-1800 dataset to show its expansibility to large-scale scenarios.

\smallskip
\noindent
\textbf{Datasets.} We re-split the Visual Genome dataset to create a VG-1800 benchmark, which contains 70,098 object categories and 1,807 predicate categories.
Different from previous large-scale VG split~\cite{zhang2019large,abdelkarim2020long}, we clean the misspellings and unreasonable relations manually and make sure all 1,807 predicate categories appear on both training and test set. For each predicate category, there are over 5 samples on the test set to provide a reliable evaluation.
Detailed statistics of VG-1800 dataset are provided in Appendix.

\smallskip
\noindent
\textbf{Tasks.}
In this work, we mainly focus on the predicate-level recognition ability and thus compare models on PREDCLS in the main paper.
For SGCLS results, please refer to the Appendix.

\smallskip
\noindent
\textbf{Metrics.}
Following \cite{zhang2019large,abdelkarim2020long}, we use accuracy (\textbf{Acc}) and mean accuracy upon all predicate classes (\textbf{mAcc}).
Similar to VG-50, the harmonic average of two metrics is reported as \textbf{F-Acc}.
In addition, we also report the number of predicate classes that the model can make at least one correct prediction, denoted as \textbf{Non-Zero}.

\smallskip
\noindent
\textbf{Baselines.}
We also include model-agnostic baselines including Focal Loss~\cite{lin2017focal}, TDE~\cite{tang2020unbiased}, and RelMix~\cite{abdelkarim2020long}, and a specific model BGNN~\cite{li2021bipartite}.

\smallskip
\noindent
\textbf{Implementation Details.}
Please refer to the Appendix for details.

\begin{table*}
    \caption{Performance of our method and baselines on VG-1800 dataset. \textbf{IETrans} denotes the Motif~\cite{zellers2018motif} model trained using our IETrans. To better compare with baselines, we show different Acc and mAcc trade-offs by setting different $k_I$.}
    \begin{center}
    \small
    \resizebox{\linewidth}{!}{%
    \begin{tabular}{ll cccc cccc cccc}
    \toprule
    &  \multirow{2}{*}{Models} & \multicolumn{4}{c}{Top-1}& \multicolumn{4}{c}{Top-5} & \multicolumn{4}{c}{Top-10}\\
    \cmidrule(lr){3-6} \cmidrule(lr){7-10} \cmidrule(lr){11-14}
    & & Acc & mAcc & F-Acc & Non-Zero & Acc & mAcc & F-Acc & Non-Zero & Acc & mAcc & F-Acc & Non-Zero \\
    \midrule
    
  &  BGNN~\cite{li2021bipartite} & \textbf{61.55} & 0.59 & 1.16 & 37 & \textbf{85.64} & 2.33 & 4.5 & 111 & \textbf{90.07} & 3.91 & 7.50 & 139  \\
    \midrule
 
 &  Motif~\cite{zellers2018motif} & 59.63 & 0.61 & 1.21 & 47 & 84.82 & 2.68 & 5.20 & 112 & 89.44 & 4.37 & 8.33 & 139  \\
 & \quad -Focal Loss & 54.65 & 0.26 & 0.52 & 14 & 79.69 & 0.79 & 1.56 & 27 & 85.21 & 1.36 & 2.68 & 41  \\
 & \quad -TDE~\cite{tang2020unbiased} & 60.00 & 0.62 & 1.23 & 45 & 85.29 & 2.77 & 5.37 & 119 & 89.92 & 4.65 & 8.84 & 152  \\
 & \quad  -RelMix~\cite{abdelkarim2020long} & 60.16 & 0.81 & 1.60 & 65 & 85.31 & 3.27 & 6.30 & 134 & 89.91 & 5.17 & 9.78 & 177  \\
 & \quad -\textbf{IETrans} $\boldsymbol{(k_I=10\%)}$ \textbf{(ours)} & 56.66 & 1.89 & 3.66 & 202 & 83.99 & 8.23 & 14.99 & 419 & 89.71 & 13.06 & 22.80 & 530 \\ 
 & \quad  -\textbf{IETrans} $\boldsymbol{(k_I=90\%)}$ \textbf{(ours)} & 27.40 & \textbf{4.70} & \textbf{8.02} & \textbf{467} & 72.48 & \textbf{13.34} & \textbf{22.53} & \textbf{741} & 83.50 & \textbf{19.12} & \textbf{31.12} & \textbf{865}  \\

 \bottomrule
    \end{tabular}}
    \end{center}
    \label{table:vg1800}
\end{table*}

\smallskip
\noindent
\textbf{Comparison with SOTAs.}
Performance of our method and baselines are shown in Table~\ref{table:vg1800}.
Based on the observation of experimental results, we have summarized the following conclusions:

\begin{wrapfigure}{R}{0.35\textwidth}
        \centering
        \includegraphics[width=1\linewidth]{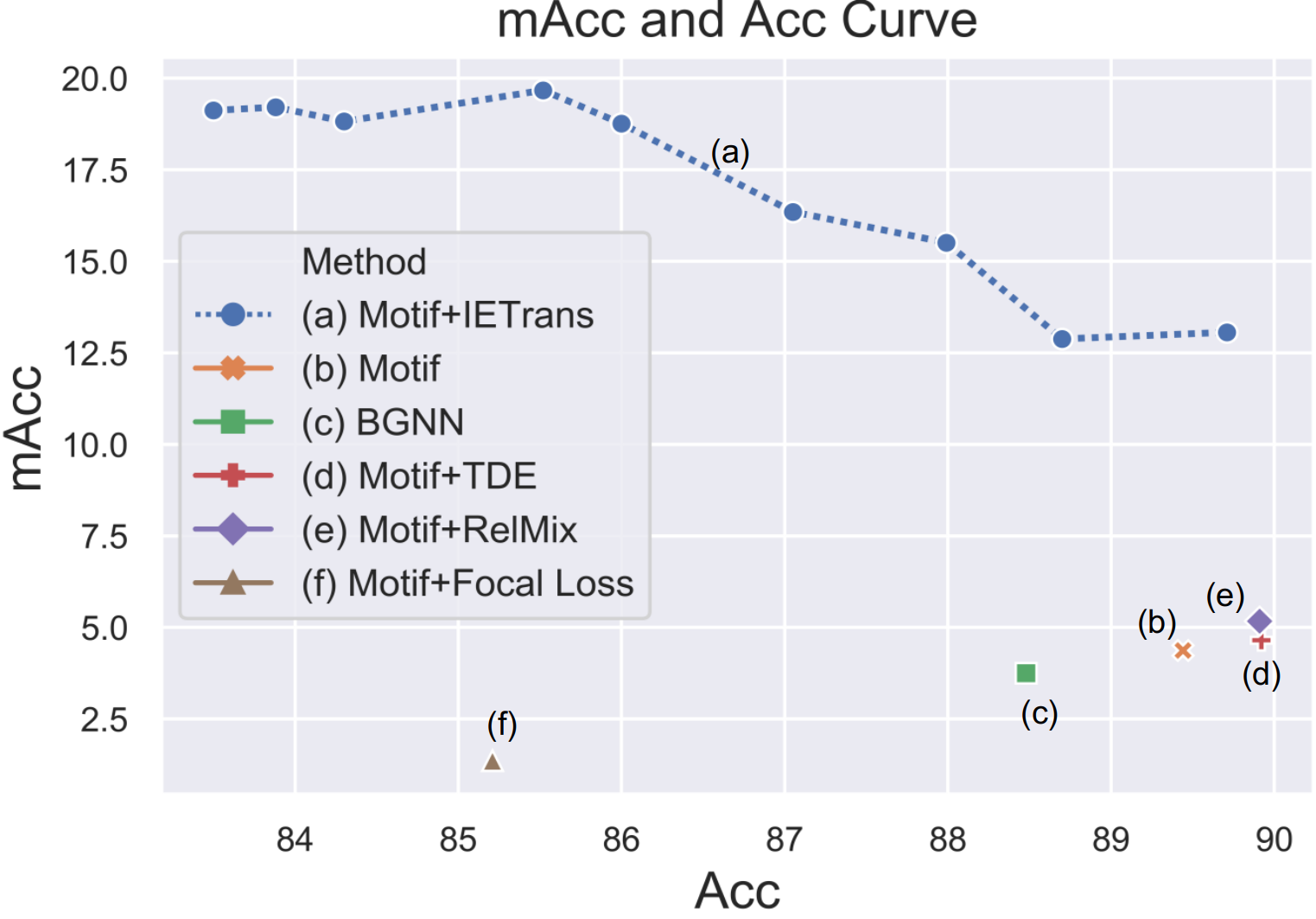}
        \caption{The mAcc and Acc curve. (a) is our IETrans method. $k_I$ is tuned to generate the blue curve. (b-f) are baselines.}
        \label{fig:abl_pr_baseline}
\end{wrapfigure}

\textbf{Our model can successfully work on large-scale settings.}
On VG-1800 dataset, the long-tail problem is even exacerbated, where hundreds of predicate classes have only less than 10 samples.
Simply increasing loss weight (Focal Loss) on tail classes can not work well.
Different from these methods, our IETrans can successfully boost the performance on mAcc while keeping competitive results on Acc.
For quantitative comparison, our \textbf{IETrans} $\boldsymbol{(k_I=10\%)}$ can significantly improve the performance on top-10 mAcc (\eg, 19.12\% vs. 4.37\%) while maintaining comparable performance on Acc.

\textbf{Compared with different baselines, our method can outperform them for overall evaluation.}
As shown in Table~\ref{table:vg1800}, our \textbf{IETrans} $\boldsymbol{(k_I=90\%)}$ can achieve best performance on F-Acc, which is over 3 times of the second highest baseline, RelMix, a method specifically designed for large-scale SGG.
To make the visualized comparison, we plot a curve of IETrans with different Acc and mAcc trade-offs by tuning $k_I$, and show the performance of other baselines as points.
As shown in Figure~\ref{fig:abl_pr_baseline}, all baselines drawn as points are under our curve, which means our method can achieve better performance than them.
Moreover, our IETrans ($k_I=90\%$) can make correct predictions on 467 predicate classes for top-1 results, while the \textbf{Non-Zero} value of all other baselines are less than 70.

\begin{figure}[t]
    \centering
    \includegraphics[width=1.\linewidth]{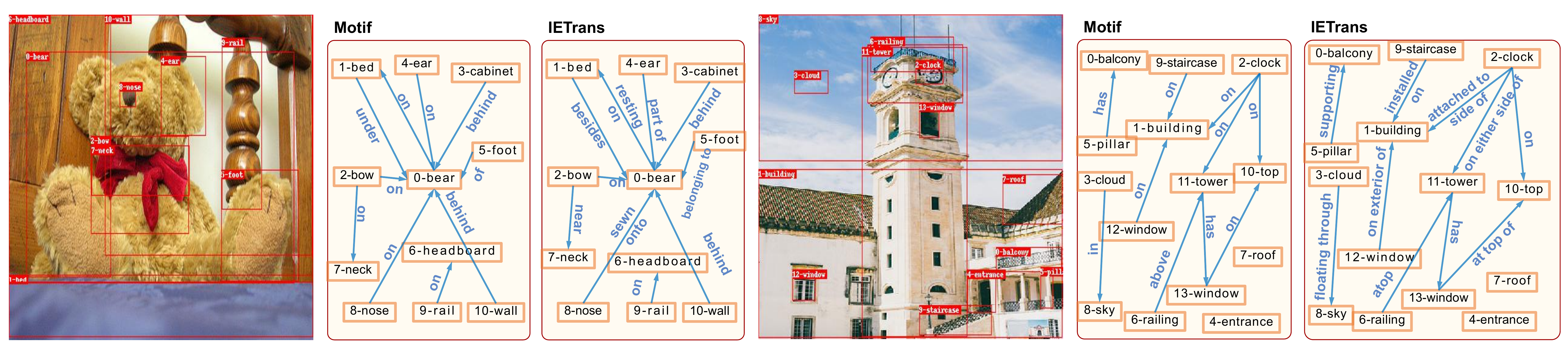}
    \caption{Visualization of raw Motif model and Motif equipped with our IETrans.}
    \label{fig:case_method}
\end{figure}

\smallskip
\noindent
\textbf{Case Studies.}
To show the potential of our method for real-world application, we provide some cases in Figure~\ref{fig:case_method}.
We can observe that our IETrans can help to generate more informative predicate classes while keeping faithful to the image content.
For example, when the Motif model only predict relational triplets like (\textit{foot}, \texttt{of}, \textit{bear}), (\textit{nose}, \texttt{on}, \textit{bear}) and (\textit{cloud}, \texttt{in}, \textit{sky}), our IETrans can generate more informative ones as (\textit{foot}, \texttt{belonging to}, \textit{bear}), (\textit{nose}, \texttt{sewn onto}, \textit{bear}), and (\textit{cloud}, \texttt{floating through}, \textit{sky}).

\subsection{Ablation Studies}
In this part, we analyse the influence of internal transfer, external transfer, and corresponding parameters, $k_I$ and $k_E$.

\begin{wrapfigure}{R}{0.35\textwidth}
        \centering
        \includegraphics[width=1\linewidth]{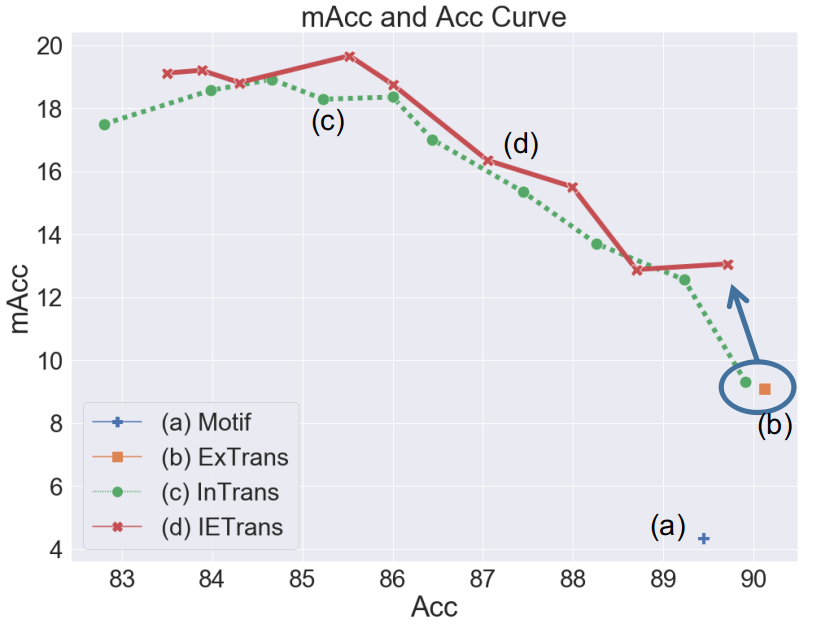}
        \caption{The mAcc and Acc curve. (a) Normally trained Motif. (b) ExTrans: external transfer. (c) InTrans: internal transfer. (d) Our proposed IETrans. $k_I$ is tuned to generate a curve. The blue circle and arrow mean that combining ExTrans and InTrans can lead to the pointed result.}
        \label{fig:abl_pr}
\end{wrapfigure}

\smallskip
\noindent
\textbf{Influence of Internal Transfer.} As shown in Figure~\ref{fig:abl_pr}, only using external transfer (yellow cube) is hard to boost the mAcc performance as much as IETrans.
The reason is that although introducing samples for tail classes, they will still be suppressed by corresponding general ones.
However, by introducing internal transfer (green point) to cope with semantic ambiguity problem, the performance (red cross) can be improved significantly on mAcc, together with minor performance drop on Acc.

\smallskip
\noindent
\textbf{Influence of External Transfer.} 
Although internal transfer can achieve huge improvement on mAcc compared with Motif, its performance is poor compared with IETrans, which shows the importance of further introducing training data by external transfer. Integration of two methods can maximize the advantages of data transfer.

\begin{figure}[t]
    \centering
    \includegraphics[width=0.7\linewidth]{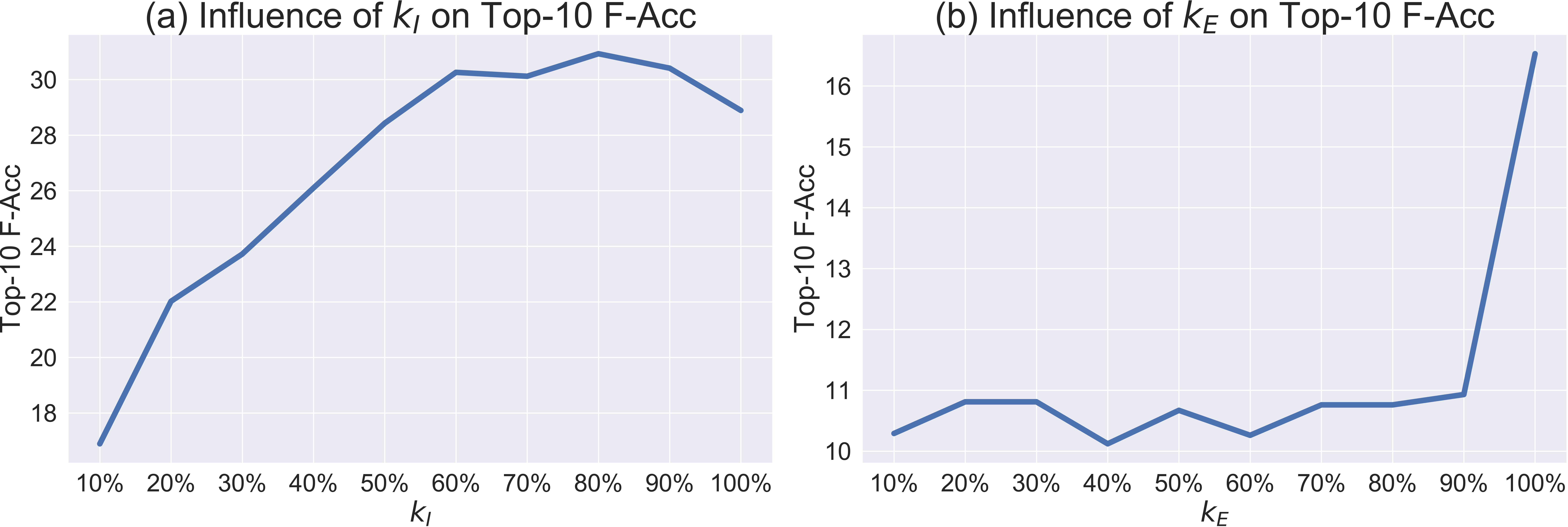}
    \caption{(a) The influence of $k_I$ in the Top-10 F-Acc with only internal transfer. (b) The influence of $k_E$ in the Top-10 F-Acc with only external transfer.}
    \label{fig:abl_k}
\end{figure}

\smallskip
\noindent
\textbf{Influence of }$ \boldsymbol{k_I} $\textbf{.}
As shown in Figure~\ref{fig:abl_k} (a), with the increase of $k_I$, the top-10 F-Acc will increase until $k_I=80\%$, and begin to decrease when $k_I>80\%$.
The phenomenon indicates that a large number of general predicates can be interpreted as informative ones.
Moving these predicates to informative ones will boost the overall performance.
However, there also exists some predicates that can not be interpreted as informative ones or be modified suitably by current methods, which is harmful to the performance of models.

\smallskip
\noindent
\textbf{Influence of }$ \boldsymbol{k_E} $\textbf{.}
As shown in Figure~\ref{fig:abl_k} (b), the overall performance increases slowly with the initial 90\% transferred data, but improves significantly with the rest 10\%. 
Note that, the data is ranked according to the \texttt{NA} score, which means that the last 10\% data is actually what the model considered as most likely to be truly negative.
The phenomenon indicates that the model may easily classify tail classes as negative samples, while this part of data is of vital significance for improving the model's ability of making informative predictions.

\subsection{Analysis of Enhanced Dataset}

\begin{figure}[t]
    \centering
    \includegraphics[width=1.0\linewidth]{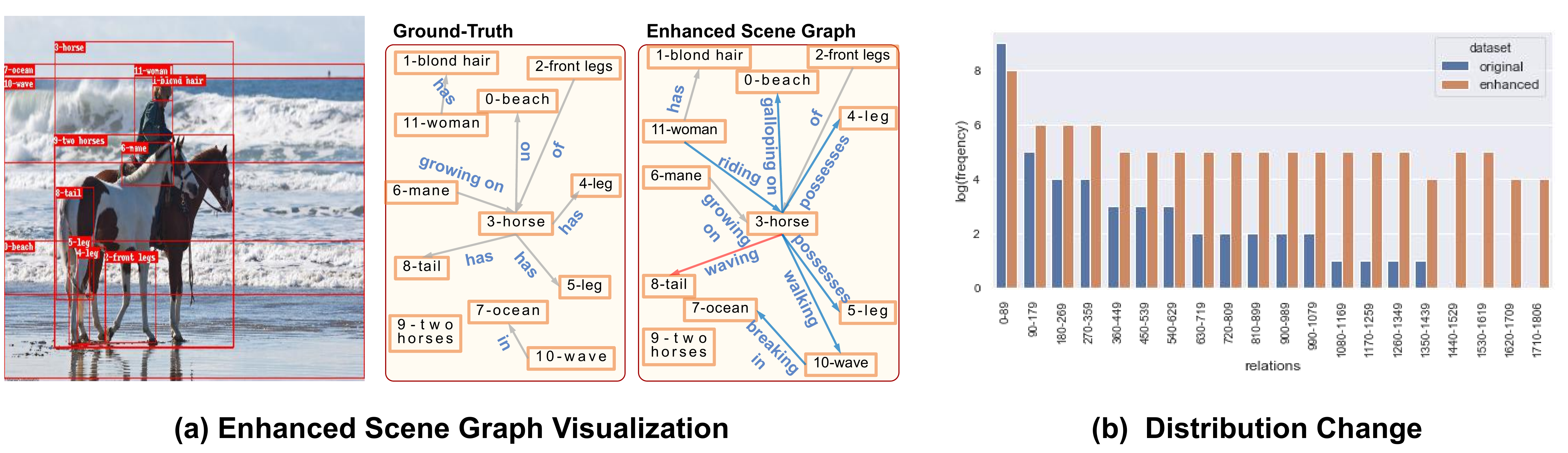}
    \caption{(a) \textbf{Enhanced Scene Graph Visualization.} Gray line denotes unchanged relation. Blue line denotes changed and reasonable relation. Red line denotes changed but unreasonable relation. (b) \textbf{Distribution Change.} The comparison between distributions of original dataset and enhanced dataset for VG-1800. 
    The x-axis is the relation id intervals from head to tail classes. The y-axis is the corresponding log-frequency.}
    \label{fig:enhanced_dataset}
\end{figure}

\smallskip
\noindent
\textbf{Enhanced Scene Graph Correctness.}
We investigate the correctness of enhanced scene graphs from the instance level.
An example is shown in Figure~\ref{fig:enhanced_dataset}(a).
We can see that the IETrans is less accurate on VG-1800, which indicates that it is more challenging to conduct precise data transfer on VG-1800.

\smallskip
\noindent
\textbf{Distribution Change.}
The distribution change is shown in Figure~\ref{fig:enhanced_dataset}(b).
We can see that our IETrans can effectively supply samples for non-head classes.


\section{Conclusion}
In this paper, we design a data transfer method named IETrans to generate an enhanced dataset for the SGG. 
The proposed IETrans consists of an internal transfer module to relabel general predicate classes as informative ones and an external transfer module to complete missed annotations.
Comprehensive experiments are conducted to show the effectiveness of our method.
In the future, we hope to extend our method to other large-scale visual recognition problems (\eg, image classification, semantic segmentation) with similar challenges.

\bigskip
\noindent
\textbf{Acknowledgements.}
This research is funded by Sea-NExT Joint Lab, Singapore. The research is also supported by the National Key Research and Development Program of China (No. 2020AAA0106500).

%
%
\bibliographystyle{splncs04}
\bibliography{egbib}

\clearpage

\appendix

\section{VG-1800 Dataset}
The VG-1800 dataset aims to provide reliable evaluation for the large-scale scene graph generation.

\subsection{Dataset Construction}
We construct the dataset based on original Visual Genome dataset~\cite{krishna2017visual} by the following steps: (1) \textbf{Filtration.} Instead of simply auto-filtering~\cite{zhang2019large} and choosing the top frequent predicate categories~\cite{abdelkarim2020long}, we manually filter out unreasonable predicate categories, including misspelling predicates (e.g., \texttt{i frot of}), adjectives (e.g., \texttt{white}), nouns (e.g., \texttt{car}), and relative clauses (e.g., \texttt{who has}).
To provide enough relation instances for robust evaluation, we retain all object categories and predicate categories with over 5 samples.
(2) \textbf{Split.} 
We split the VG dataset into $70$\% training and $30$\% test.
Following VG-50 split~\cite{xu2017scene}, we further split out $5,000$ images from the training set as the validation set, and ensure at least $5$ samples on the test set and at least $1$ samples on the training set for each predicate category.

\subsection{Dataset Statistics}
Finally, the dataset contains $70,098$ object categories, $1,807$ predicate categories and $272,084$ distinct relation triplets.
It consists of $66,289$, $4,995$, and $32,893$ images for training set, validation set and test set respectively.
There are on average $19.5$ objects and $16.0$ relations for each image.

\begin{figure*}
    \centering
    \includegraphics[width=1\linewidth]{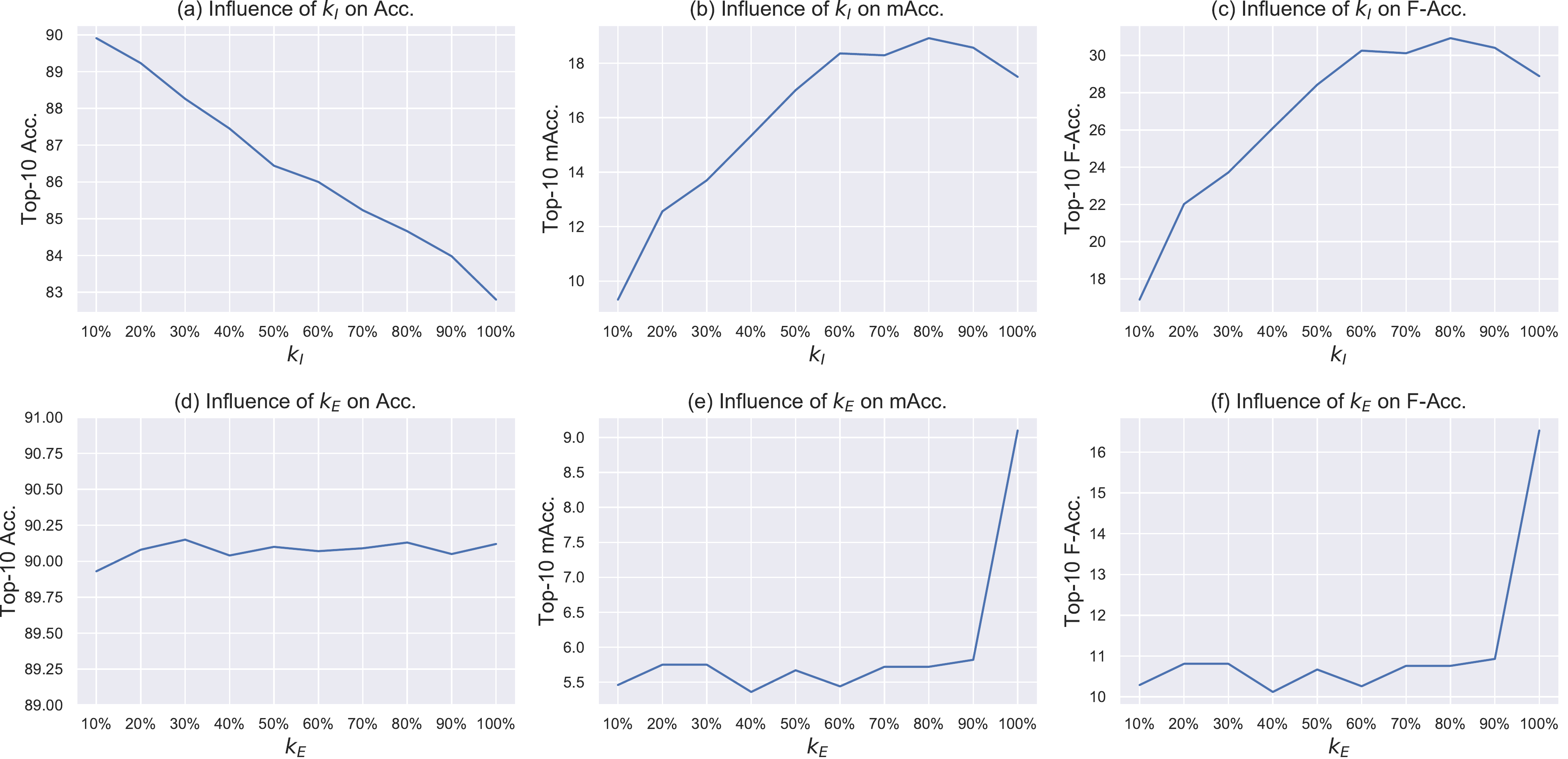}
    \caption{Influence of $k_I$ and $k_E$ in different metrics.}
    \label{fig:abl_kIE}
\end{figure*}

\begin{figure}[t]
    \centering
    \includegraphics[width=0.8\linewidth]{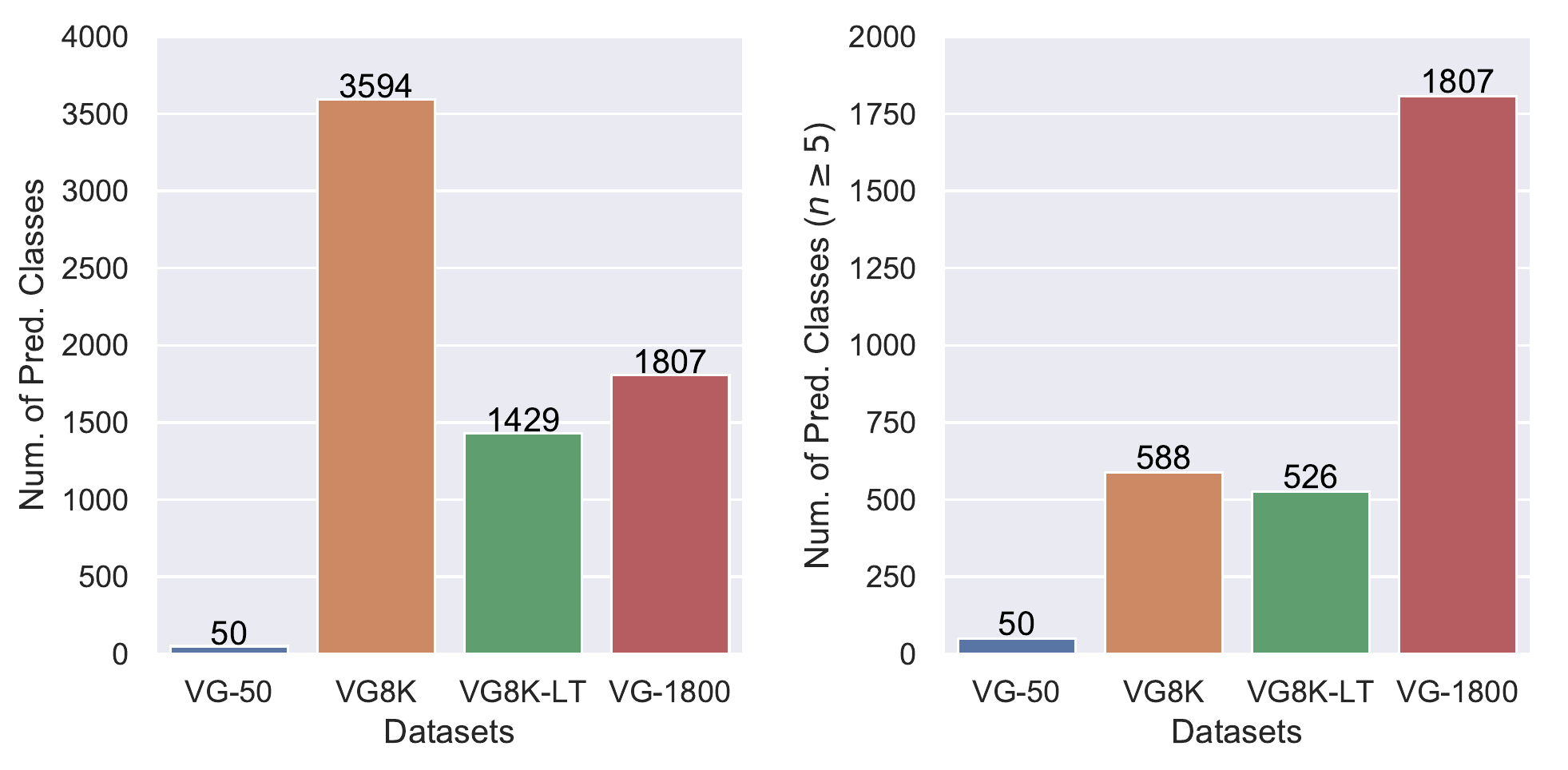}
    \caption{Comparison of number of predicate classes between different splits on test set. The $n \geq 5$ denotes predicate classes having no less than 5 samples.}
    \label{fig:data_cmp}
\end{figure}

\smallskip
\noindent
\textbf{Comparison with Other VG Splits.}
We also compare our VG-1800 split with other splits based on Visual Genome~\cite{krishna2017visual} dataset, including a conventional VG-50 and the other two large-scale SGG splits VG8K and VG8K-LT.
Our VG-1800 can provide a more reliable evaluation for large-scale SGG.
(1) When compared with VG8K, we provide a much cleaner dataset by manually cleaning the noise.
For example, VG8K does not filter out nouns and adjectives, which will lead to an unreliable evaluation.
(2) When compared with VG8K-LT, a cleaner version of VG8K, we provide a much stable evaluation for large amount of tail classes.
More specifically, as shown in Table~\ref{tab:data_cmp}, our VG-1800 contains more test images.
Meanwhile, VG-1800 also contains more samples of tail classes.
As shown in Figure~\ref{fig:data_cmp}, our VG-1800 has 1,807 predicate classes with no less than 5 samples, while VG8K-LT has only 526 classes that have no less than 5 samples.

\begin{table}[]
    \setlength{\tabcolsep}{15pt}
    \centering
    \caption{Comparison between different datasets' predicate filtration and split ratio. Filtration denotes the method to remove noisy predicates. The Train, Val, and Test denote number of images in training set, validation set and test set.}
    \label{tab:data_cmp}
    \resizebox{0.9\textwidth}{!}{
    \begin{tabular}{l c c c c}
    \toprule
    Dataset & Filtration & Train & Val & Test \\
    \midrule
    VG-50~\cite{xu2017scene} & - & 57,723 & 5,000 & 26,446 \\
    VG8K~\cite{zhang2019large} & Auto & 97,961 & 2,000 & 4,871 \\
    VG8K-LT~\cite{abdelkarim2020long} & Auto & 97,623 & 1,999 & 4,860 \\
    VG-1800 & Manual & 66,289 & 4,995 & 32,893 \\

    \bottomrule
    \end{tabular}}

\end{table}

\section{Implementation Details}
\subsection{VG-50}
All normally trained baseline models including Motif, Transformer, VCTree, and GPS-Net are reproduced by us.
For fair comparison, we equally remove all resampling and reweighting strategies.
Morever, to encourage informative SGG, we remove the frequency bias in the training and inference of base models, which may make the result tend to have higher mR@K, F@K and lower R@K than results in their original papers.

For Transformer, some implementation details are different. 
The batch size can be enlarged to 16 on 2 GPUs.
The learning rate is reduced to 0.08 for PREDCLS and SGDET for training stability.
For Transformer on SGCLS, where the training is even more unstable, we further lower the learning rate to 0.016.
All experiments are done on RTX-2080ti GPUs. 

\subsection{VG-1800}
Compared with VG-50, the same backbone, parameter fixation, learning rate, optimizer, and learning schedule are used on VG-1800 dataset.
Specially, due to the significant increase of (\textit{subject}, \texttt{predicate}, \textit{object}) combinations, we equally remove all frequency bias items on VG-1800 to reduce machines' memory usage.
For internal and external transfer, the $k_I$ is set to 90\% and $k_E$ is set to 100\%.

For baselines, we find that due to the significant difference between the number of head predicates and tail predicates, the bi-level resampling in BGNN~\cite{li2021bipartite} will make the model pay most of the attention on tail classes while ignoring head classes.
The drop out rate of images that do not contain rare predicates are set to almost 100\%.
This lead to a bad convergence of BGNN.
Thus, we remove bi-level resampling for BGNN results.
For RelMix~\cite{abdelkarim2020long}, we equip the proposed VilHub loss and predicate feature mixup to the Neural Motif model.
Similar with bi-level sampling, we find that the reweighting strategy also lead to worse results.
Thus, we do not include a reweighting version like VG-50.


\section{Supplementary Experiments}
\subsection{Influence of $\boldsymbol{k_I}$ abd $\boldsymbol{k_E}$}
\smallskip
\noindent
\textbf{Influence of } $\boldsymbol{k_I}$\textbf{.}
To provide a more detailed analysis on the influence of $k_I$, we report the performance on Acc and mAcc with different $k_I$.
As shown in Figure~\ref{fig:abl_kIE}, with the increase of internal transfer percentage $k_I$, Acc decreases linearly, while mAcc first increases when $k_I\leq80\%$ and then decreases.
The phenomenon shows that transferring more in internal transfer does not necessarily mean higher mAcc.
For VG-1800, The first 80\% internal data transfer is helpful to improve mAcc, while the last confident 20\% will harm the overall performance.
We guess the last 20\% data may contain too noise, which will lower the data quality for model training.

\smallskip
\noindent
\textbf{Influence of } $\boldsymbol{k_E}$\textbf{.}
As for $k_E$, external transfer shows almost no influence on Acc and mAcc when $k_E \leq 90\%$, while significantly boost mAcc when $k_E=100\%$.
Contrary to our observations for $k_I$, the last 10\% samples which are believed to be unuseful by models, seem to bring the most profitable boost for mAcc.
We guess the reason is that model can not distinguish well between tail classes and \texttt{NA} samples, while this part of the data is essential to provide more training samples for tail classes.
\subsection{Adaptive Threshold.}
In our IETrans, when determining how much data to transfer, we equally use a fixed percentage number for all relational triplets, which seems to be sub-optimal.
Thus, we also tried an adaptive threshold by considering the prediction score of concrete relational triplet instances.

For internal transfer, given a general relational triplet instance $(o_s, p_G, o_o)$, we are required to decide whether to transfer to its corresponding informative type $p_I$.
We denote the model's prediction score of object pair $(o_s, o_o)$  on $p_I$ as $s_{p_I}^*$.
We denote the average and standard error of all $(c_{o_s}, p_I, c_{o_o})$ relational triplet instances' prediction score on $p_I$ as $\mu_{I}$ and $\sigma_{I}$.
We conduct the transfer when $s_{p_I}^*$ satisfies:
\begin{equation}
    s_{p_I}^* > \mu_{I} +k \sigma_{I},
\end{equation}
where $k$ is a hyperparameter.
The intuition is that if a general instance $(o_s, p_G, o_o)$'s prediction score is over the average of all real $(c_{o_s}, p_I, c_{o_o})$'s prediction scores on $p_I$,  $(o_s, p_G, o_o)$ can probably be relabeled as an informative one.
Meanwhile, the standard error is considered to further control the adaptive threshold.

\begin{figure}[t]
    \centering
    \includegraphics[width=0.5\linewidth]{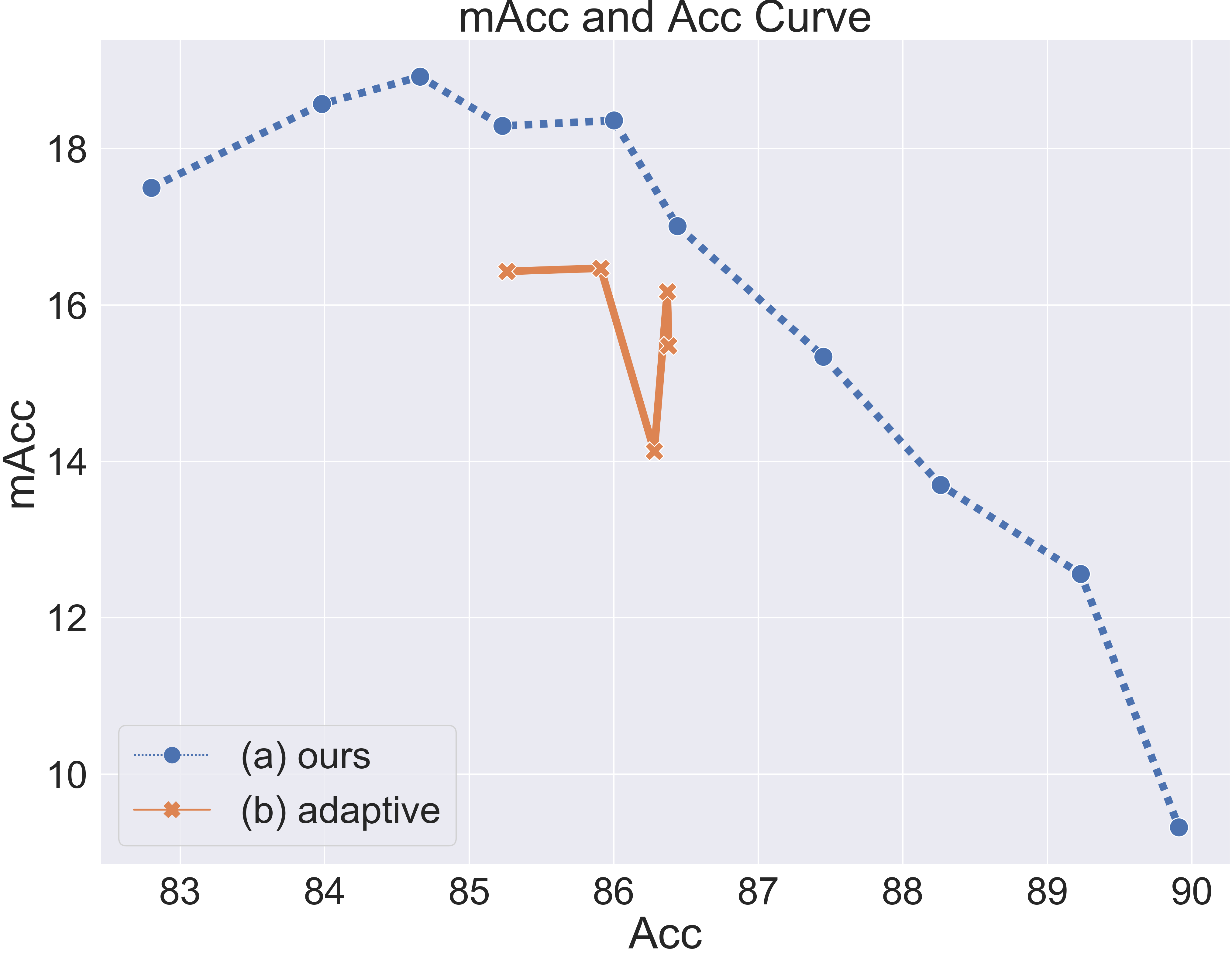}
    \caption{Comparison of adaptive thresholds and fixed percentages for Internal Transfer.}
    \label{fig:data_adp}
\end{figure}

By choosing different $k$ including $\{-1.0, -0.5, 0.0, 0.5, 1.0\}$, we can get a curve with different Acc and mAcc trade-offs.
As shown in Figure~\ref{fig:data_adp}, the model with adaptive thresholds is overall worse than our fixed percentage.

A possible explanation is that the prediction score of an instance is non-linearly dependent on the number of its own instances and its similarity with different general classes, which results in inconsistency among different relational triplets.
Especially when the number of an informative relational triplet is very small, the average of its prediction score is often near to zero, which will easily lead to an over-transfer problem.
Thus, we leave the design of a more intelligent adaptive threshold for future work.

For external transfer, as shown in the paper, the prediction scores on \texttt{NA} of missed annotated samples are unreliable, i.e. the samples with the highest \texttt{NA} score bring maximum benefits for the model's performance. 

\smallskip
\noindent
\textbf{SGCLS Results on VG-1800.}
\begin{table*}
    \caption{SGCLS triplet-level evaluation results on VG-1800 dataset.}
    \begin{center}
    \small
    \resizebox{\linewidth}{!}{%
    \begin{tabular}{ll cccc cccc cccc}
    \toprule
    & \multirow{2}{*}{Models} & \multicolumn{4}{c}{Top-1}& \multicolumn{4}{c}{Top-5} & \multicolumn{4}{c}{Top-10}\\
    \cmidrule(lr){3-6} \cmidrule(lr){7-10} \cmidrule(lr){11-14}
    & & Acc & mAcc & F-Acc & Non-Zero & Acc & mAcc & F-Acc & Non-Zero & Acc & mAcc & F-Acc & Non-Zero \\
    \midrule
 &  BGNN~\cite{li2021bipartite} & 16.29 & 0.18 & 0.36 & 22 & 22.99 & 0.86 & 1.65 & 159 & 24.15& 1.48 & 2.78 & 221 \\
    \midrule
 &  Motif~\cite{zellers2018motif} & \textbf{18.93} & 0.18 & 0.36 & 37 & \textbf{26.08} & 0.74 & 1.43 & 90 & \textbf{27.28}& 1.15 & 2.21 & 121 \\
 & \quad -Focal Loss & 18.55 & 0.14 & 0.28 & 29 & 25.91 & 0.51 & 1.00 & 52 & 27.14 & 0.76 & 1.48 & 80  \\
 & \quad  -TDE~\cite{tang2020unbiased} & 18.02 & 0.11 & 0.22 & 15 & 24.85 & 0.38 & 0.75 & 38 & 26.14 & 0.56 & 1.10 & 53  \\
 & \quad -RelMix~\cite{abdelkarim2020long} & 18.27 & 0.22 & 0.43 & 47 & 25.57 & 0.83 & 1.60 & 100 & 26.71 & 1.26 & 2.42 & 130  \\
 & \quad -\textbf{IETrans} $\boldsymbol{(k_I=10\%)}$ \textbf{(ours)} & 18.24 & 1.68 & 3.16 & 212 & 25.80 & 1.68 & 3.16 & 212 & 27.25 & 2.54 & 4.65 & 264 \\ 
 & \quad -\textbf{IETrans} $\boldsymbol{(k_I=90\%)}$ \textbf{(ours)} & 4.91 & \textbf{1.78} & \textbf{2.62} & \textbf{298} & 20.66 & \textbf{4.72} & \textbf{7.68} & \textbf{538} & 24.85 & \textbf{6.54} & \textbf{10.36} & \textbf{637}  \\
 \bottomrule
    \end{tabular}}
    \end{center}
    \label{table:vg1800_sgcls}
\end{table*}
We also evaluate our method on SGCLS task on VG-1800 dataset.
As shown in table~\ref{table:vg1800_sgcls}, the comparison with other baselines is similar to the results on PREDCLS task.
When compared with Motif, our IETrans ($k_I=10\%$) can achieve significant improvement on top-1 mAcc and Non-Zero metrics (over 5 times of Motif) with negligible degeneration (less than 1 point) on top-1 Acc metric.
Our IETrans ($k_I=90\%$) can further boost the mAcc and Non-Zero metrics, which shows the ability of our IETrans to generate informative scene graphs.
When compared with other baselines, our IEtrans can achieve the best F-Acc metrics across top-1, top-5, and top-10 evaluations.
However, there is a large gap between SGCLS results and PREDCLS results (\eg, 2.62\% vs. 4.70\% for top-1 F-Acc of IETrans ($k_I=90\%$)), which indicates that further effort should be made to explore the joint optimization of both objects and predicates.

\section{Discovered Visual Hierarchy Analysis}

\smallskip
\noindent
\textbf{Visual Hierarchy Evaluation.} A key element of conducting correct internal transfer is to find reasonable general-informative relation pairs.
To evaluate the precision, we randomly choose 50 pairs with over 3 samples being transferred, so as to avoid involving too many noise-to-noise pairs.
Then, human evaluation is conducted.
The ratio of reasonable general-informative pairs is \textbf{76\%} for VG-50, and \textbf{74\%} for VG-1800.

\smallskip
\noindent
\textbf{Visualization.} In the following, we show 100 discovered general-informative pairs for both VG-50 and VG-1800.
The pairs are ranked by the number of samples which are transferred.


{
\begin{longtable}{ l }
\caption{Examples of discovered visual hierarchy in VG-50}
\label{table:vg50_vis_h}\\
\hline
(\textit{window}, \texttt{on}, \textit{building}) $\rightarrow$ (\textit{window}, \texttt{part of}, \textit{building}) \\ \hline
(\textit{man}, \texttt{wearing}, \textit{arm}) $\rightarrow$ (\textit{man}, \texttt{wears}, \textit{arm}) \\ \hline
(\textit{boy}, \texttt{wearing}, \textit{boy}) $\rightarrow$ (\textit{boy}, \texttt{wears}, \textit{boy}) \\ \hline
(\textit{pillow}, \texttt{on}, \textit{bed}) $\rightarrow$ (\textit{pillow}, \texttt{lying on}, \textit{bed}) \\ \hline
(\textit{building}, \texttt{has}, \textit{building}) $\rightarrow$ (\textit{building}, \texttt{made of}, \textit{building}) \\ \hline
(\textit{sign}, \texttt{on}, \textit{building}) $\rightarrow$ (\textit{sign}, \texttt{mounted on}, \textit{building}) \\ \hline
(\textit{arm}, \texttt{of}, \textit{arm}) $\rightarrow$ (\textit{arm}, \texttt{belonging to}, \textit{arm}) \\ \hline
(\textit{sign}, \texttt{on}, \textit{building}) $\rightarrow$ (\textit{sign}, \texttt{hanging from}, \textit{building}) \\ \hline
(\textit{man}, \texttt{wearing}, \textit{bag}) $\rightarrow$ (\textit{man}, \texttt{wears}, \textit{bag}) \\ \hline
(\textit{window}, \texttt{on}, \textit{building}) $\rightarrow$ (\textit{window}, \texttt{belonging to}, \textit{building}) \\ \hline
(\textit{car}, \texttt{on}, \textit{building}) $\rightarrow$ (\textit{car}, \texttt{parked on}, \textit{building}) \\ \hline
(\textit{clock}, \texttt{on}, \textit{building}) $\rightarrow$ (\textit{clock}, \texttt{mounted on}, \textit{building}) \\ \hline
(\textit{man}, \texttt{wearing}, \textit{building}) $\rightarrow$ (\textit{man}, \texttt{wears}, \textit{building}) \\ \hline
(\textit{man}, \texttt{has}, \textit{arm}) $\rightarrow$ (\textit{man}, \texttt{wears}, \textit{arm}) \\ \hline
(\textit{man}, \texttt{wearing}, \textit{boot}) $\rightarrow$ (\textit{man}, \texttt{wears}, \textit{boot}) \\ \hline
(\textit{bottle}, \texttt{on}, \textit{bottle}) $\rightarrow$ (\textit{bottle}, \texttt{sitting on}, \textit{bottle}) \\ \hline
(\textit{window}, \texttt{on}, \textit{bus}) $\rightarrow$ (\textit{window}, \texttt{belonging to}, \textit{bus}) \\ \hline
(\textit{book}, \texttt{on}, \textit{book}) $\rightarrow$ (\textit{book}, \texttt{above}, \textit{book}) \\ \hline
(\textit{man}, \texttt{has}, \textit{arm}) $\rightarrow$ (\textit{man}, \texttt{with}, \textit{arm}) \\ \hline
(\textit{ear}, \texttt{of}, \textit{ear}) $\rightarrow$ (\textit{ear}, \texttt{belonging to}, \textit{ear}) \\ \hline
(\textit{hand}, \texttt{of}, \textit{arm}) $\rightarrow$ (\textit{hand}, \texttt{belonging to}, \textit{arm}) \\ \hline
(\textit{bottle}, \texttt{on}, \textit{bottle}) $\rightarrow$ (\textit{bottle}, \texttt{above}, \textit{bottle}) \\ \hline
(\textit{window}, \texttt{in}, \textit{building}) $\rightarrow$ (\textit{window}, \texttt{part of}, \textit{building}) \\ \hline
(\textit{light}, \texttt{on}, \textit{building}) $\rightarrow$ (\textit{light}, \texttt{mounted on}, \textit{building}) \\ \hline
(\textit{door}, \texttt{on}, \textit{building}) $\rightarrow$ (\textit{door}, \texttt{to}, \textit{building}) \\ \hline
(\textit{food}, \texttt{on}, \textit{food}) $\rightarrow$ (\textit{food}, \texttt{lying on}, \textit{food}) \\ \hline
(\textit{bowl}, \texttt{on}, \textit{bowl}) $\rightarrow$ (\textit{bowl}, \texttt{above}, \textit{bowl}) \\ \hline
(\textit{man}, \texttt{wearing}, \textit{man}) $\rightarrow$ (\textit{man}, \texttt{wears}, \textit{man}) \\ \hline
(\textit{flower}, \texttt{in}, \textit{flower}) $\rightarrow$ (\textit{flower}, \texttt{painted on}, \textit{flower}) \\ \hline
(\textit{woman}, \texttt{wearing}, \textit{bag}) $\rightarrow$ (\textit{woman}, \texttt{wears}, \textit{bag}) \\ \hline
(\textit{tree}, \texttt{near}, \textit{building}) $\rightarrow$ (\textit{tree}, \texttt{in front of}, \textit{building}) \\ \hline
(\textit{woman}, \texttt{wearing}, \textit{arm}) $\rightarrow$ (\textit{woman}, \texttt{wears}, \textit{arm}) \\ \hline
(\textit{window}, \texttt{of}, \textit{building}) $\rightarrow$ (\textit{window}, \texttt{part of}, \textit{building}) \\ \hline
(\textit{clock}, \texttt{on}, \textit{building}) $\rightarrow$ (\textit{clock}, \texttt{part of}, \textit{building}) \\ \hline
(\textit{bus}, \texttt{on}, \textit{building}) $\rightarrow$ (\textit{bus}, \texttt{parked on}, \textit{building}) \\ \hline
(\textit{man}, \texttt{wearing}, \textit{coat}) $\rightarrow$ (\textit{man}, \texttt{wears}, \textit{coat}) \\ \hline
(\textit{building}, \texttt{has}, \textit{building}) $\rightarrow$ (\textit{building}, \texttt{with}, \textit{building}) \\ \hline
(\textit{boy}, \texttt{wearing}, \textit{arm}) $\rightarrow$ (\textit{boy}, \texttt{wears}, \textit{arm}) \\ \hline
(\textit{man}, \texttt{on}, \textit{arm}) $\rightarrow$ (\textit{man}, \texttt{riding}, \textit{arm}) \\ \hline
(\textit{tree}, \texttt{has}, \textit{branch}) $\rightarrow$ (\textit{tree}, \texttt{with}, \textit{branch}) \\ \hline
(\textit{woman}, \texttt{wearing}, \textit{boot}) $\rightarrow$ (\textit{woman}, \texttt{wears}, \textit{boot}) \\ \hline
(\textit{pillow}, \texttt{on}, \textit{bed}) $\rightarrow$ (\textit{pillow}, \texttt{above}, \textit{bed}) \\ \hline
(\textit{woman}, \texttt{has}, \textit{arm}) $\rightarrow$ (\textit{woman}, \texttt{with}, \textit{arm}) \\ \hline
(\textit{window}, \texttt{on}, \textit{building}) $\rightarrow$ (\textit{window}, \texttt{to}, \textit{building}) \\ \hline
(\textit{glass}, \texttt{on}, \textit{bottle}) $\rightarrow$ (\textit{glass}, \texttt{sitting on}, \textit{bottle}) \\ \hline
(\textit{sign}, \texttt{on}, \textit{building}) $\rightarrow$ (\textit{sign}, \texttt{says}, \textit{building}) \\ \hline
(\textit{man}, \texttt{wearing}, \textit{bike}) $\rightarrow$ (\textit{man}, \texttt{wears}, \textit{bike}) \\ \hline
(\textit{tire}, \texttt{on}, \textit{building}) $\rightarrow$ (\textit{tire}, \texttt{on back of}, \textit{building}) \\ \hline
(\textit{branch}, \texttt{on}, \textit{branch}) $\rightarrow$ (\textit{branch}, \texttt{growing on}, \textit{branch}) \\ \hline
(\textit{book}, \texttt{on}, \textit{book}) $\rightarrow$ (\textit{book}, \texttt{laying on}, \textit{book}) \\ \hline
(\textit{car}, \texttt{on}, \textit{car}) $\rightarrow$ (\textit{car}, \texttt{parked on}, \textit{car}) \\ \hline
(\textit{man}, \texttt{wearing}, \textit{bench}) $\rightarrow$ (\textit{man}, \texttt{wears}, \textit{bench}) \\ \hline
(\textit{elephant}, \texttt{has}, \textit{ear}) $\rightarrow$ (\textit{elephant}, \texttt{using}, \textit{ear}) \\ \hline
(\textit{person}, \texttt{on}, \textit{beach}) $\rightarrow$ (\textit{person}, \texttt{standing on}, \textit{beach}) \\ \hline
(\textit{wing}, \texttt{on}, \textit{plane}) $\rightarrow$ (\textit{wing}, \texttt{attached to}, \textit{plane}) \\ \hline
(\textit{windshield}, \texttt{on}, \textit{building}) $\rightarrow$ (\textit{windshield}, \texttt{of}, \textit{building}) \\ \hline
(\textit{arm}, \texttt{on}, \textit{arm}) $\rightarrow$ (\textit{arm}, \texttt{belonging to}, \textit{arm}) \\ \hline
(\textit{woman}, \texttt{holding}, \textit{bag}) $\rightarrow$ (\textit{woman}, \texttt{carrying}, \textit{bag}) \\ \hline
(\textit{window}, \texttt{on}, \textit{bike}) $\rightarrow$ (\textit{window}, \texttt{part of}, \textit{bike}) \\ \hline
(\textit{ear}, \texttt{on}, \textit{ear}) $\rightarrow$ (\textit{ear}, \texttt{belonging to}, \textit{ear}) \\ \hline
(\textit{man}, \texttt{on}, \textit{beach}) $\rightarrow$ (\textit{man}, \texttt{walking on}, \textit{beach}) \\ \hline
(\textit{boy}, \texttt{has}, \textit{boy}) $\rightarrow$ (\textit{boy}, \texttt{wears}, \textit{boy}) \\ \hline
(\textit{roof}, \texttt{on}, \textit{building}) $\rightarrow$ (\textit{roof}, \texttt{covering}, \textit{building}) \\ \hline
(\textit{leaf}, \texttt{on}, \textit{branch}) $\rightarrow$ (\textit{leaf}, \texttt{growing on}, \textit{branch}) \\ \hline
(\textit{head}, \texttt{of}, \textit{arm}) $\rightarrow$ (\textit{head}, \texttt{belonging to}, \textit{arm}) \\ \hline
(\textit{wheel}, \texttt{on}, \textit{building}) $\rightarrow$ (\textit{wheel}, \texttt{on back of}, \textit{building}) \\ \hline
(\textit{tree}, \texttt{near}, \textit{building}) $\rightarrow$ (\textit{tree}, \texttt{along}, \textit{building}) \\ \hline
(\textit{bird}, \texttt{on}, \textit{bird}) $\rightarrow$ (\textit{bird}, \texttt{sitting on}, \textit{bird}) \\ \hline
(\textit{door}, \texttt{on}, \textit{door}) $\rightarrow$ (\textit{door}, \texttt{to}, \textit{door}) \\ \hline
(\textit{woman}, \texttt{has}, \textit{bag}) $\rightarrow$ (\textit{woman}, \texttt{with}, \textit{bag}) \\ \hline
(\textit{man}, \texttt{wearing}, \textit{hat}) $\rightarrow$ (\textit{man}, \texttt{wears}, \textit{hat}) \\ \hline
(\textit{man}, \texttt{on}, \textit{arm}) $\rightarrow$ (\textit{man}, \texttt{standing on}, \textit{arm}) \\ \hline
(\textit{sign}, \texttt{on}, \textit{building}) $\rightarrow$ (\textit{sign}, \texttt{attached to}, \textit{building}) \\ \hline
(\textit{letter}, \texttt{on}, \textit{building}) $\rightarrow$ (\textit{letter}, \texttt{painted on}, \textit{building}) \\ \hline
(\textit{bird}, \texttt{on}, \textit{bird}) $\rightarrow$ (\textit{bird}, \texttt{standing on}, \textit{bird}) \\ \hline
(\textit{ear}, \texttt{of}, \textit{cat}) $\rightarrow$ (\textit{ear}, \texttt{belonging to}, \textit{cat}) \\ \hline
(\textit{window}, \texttt{on}, \textit{bench}) $\rightarrow$ (\textit{window}, \texttt{part of}, \textit{bench}) \\ \hline
(\textit{window}, \texttt{near}, \textit{building}) $\rightarrow$ (\textit{window}, \texttt{part of}, \textit{building}) \\ \hline
(\textit{wheel}, \texttt{on}, \textit{bike}) $\rightarrow$ (\textit{wheel}, \texttt{on back of}, \textit{bike}) \\ \hline
(\textit{building}, \texttt{near}, \textit{building}) $\rightarrow$ (\textit{building}, \texttt{across}, \textit{building}) \\ \hline
(\textit{elephant}, \texttt{has}, \textit{ear}) $\rightarrow$ (\textit{elephant}, \texttt{between}, \textit{ear}) \\ \hline
(\textit{man}, \texttt{has}, \textit{bag}) $\rightarrow$ (\textit{man}, \texttt{with}, \textit{bag}) \\ \hline
(\textit{engine}, \texttt{on}, \textit{plane}) $\rightarrow$ (\textit{engine}, \texttt{mounted on}, \textit{plane}) \\ \hline
(\textit{man}, \texttt{wearing}, \textit{chair}) $\rightarrow$ (\textit{man}, \texttt{wears}, \textit{chair}) \\ \hline
(\textit{woman}, \texttt{wearing}, \textit{building}) $\rightarrow$ (\textit{woman}, \texttt{wears}, \textit{building}) \\ \hline
(\textit{sign}, \texttt{on}, \textit{sign}) $\rightarrow$ (\textit{sign}, \texttt{mounted on}, \textit{sign}) \\ \hline
(\textit{plate}, \texttt{on}, \textit{bowl}) $\rightarrow$ (\textit{plate}, \texttt{above}, \textit{bowl}) \\ \hline
(\textit{man}, \texttt{wearing}, \textit{face}) $\rightarrow$ (\textit{man}, \texttt{wears}, \textit{face}) \\ \hline
(\textit{leg}, \texttt{of}, \textit{giraffe}) $\rightarrow$ (\textit{leg}, \texttt{part of}, \textit{giraffe}) \\ \hline
(\textit{pillow}, \texttt{above}, \textit{bed}) $\rightarrow$ (\textit{pillow}, \texttt{lying on}, \textit{bed}) \\ \hline
(\textit{tire}, \texttt{on}, \textit{bike}) $\rightarrow$ (\textit{tire}, \texttt{on back of}, \textit{bike}) \\ \hline
(\textit{leg}, \texttt{of}, \textit{arm}) $\rightarrow$ (\textit{leg}, \texttt{belonging to}, \textit{arm}) \\ \hline
(\textit{man}, \texttt{wearing}, \textit{cap}) $\rightarrow$ (\textit{man}, \texttt{wears}, \textit{cap}) \\ \hline
(\textit{bird}, \texttt{has}, \textit{bird}) $\rightarrow$ (\textit{bird}, \texttt{with}, \textit{bird}) \\ \hline
(\textit{trunk}, \texttt{of}, \textit{ear}) $\rightarrow$ (\textit{trunk}, \texttt{belonging to}, \textit{ear}) \\ \hline
(\textit{roof}, \texttt{of}, \textit{building}) $\rightarrow$ (\textit{roof}, \texttt{covering}, \textit{building}) \\ \hline
(\textit{plate}, \texttt{on}, \textit{bottle}) $\rightarrow$ (\textit{plate}, \texttt{above}, \textit{bottle}) \\ \hline
(\textit{man}, \texttt{wearing}, \textit{ear}) $\rightarrow$ (\textit{man}, \texttt{wears}, \textit{ear}) \\ \hline
(\textit{man}, \texttt{has}, \textit{building}) $\rightarrow$ (\textit{man}, \texttt{wears}, \textit{building}) \\ \hline
(\textit{window}, \texttt{on}, \textit{arm}) $\rightarrow$ (\textit{window}, \texttt{part of}, \textit{arm}) \\ \hline
\end{longtable}
}

{
\begin{longtable}{ l }
\caption{Examples of discovered visual hierarchy in VG-1800}
\label{table:vg1800_vis_h}\\
\hline
(\textit{window}, \texttt{on}, \textit{building}) $\rightarrow$ (\textit{window}, \texttt{on exterior of}, \textit{building}) \\ \hline
(\textit{man}, \texttt{wearing}, \textit{arm}) $\rightarrow$ (\textit{man}, \texttt{wearing striped}, \textit{arm}) \\ \hline
(\textit{arm}, \texttt{of}, \textit{arm}) $\rightarrow$ (\textit{arm}, \texttt{stretched out on}, \textit{arm}) \\ \hline
(\textit{man}, \texttt{has}, \textit{arm}) $\rightarrow$ (\textit{man}, \texttt{stretching out}, \textit{arm}) \\ \hline
(\textit{cloud}, \texttt{in}, \textit{cloud}) $\rightarrow$ (\textit{cloud}, \texttt{floating through}, \textit{cloud}) \\ \hline
(\textit{pillow}, \texttt{on}, \textit{bed}) $\rightarrow$ (\textit{pillow}, \texttt{propped up on}, \textit{bed}) \\ \hline
(\textit{tree}, \texttt{in}, \textit{background}) $\rightarrow$ (\textit{tree}, \texttt{visible in}, \textit{background}) \\ \hline
(\textit{leg}, \texttt{of}, \textit{arm}) $\rightarrow$ (\textit{leg}, \texttt{belonging to}, \textit{arm}) \\ \hline
(\textit{boat}, \texttt{in}, \textit{boat}) $\rightarrow$ (\textit{boat}, \texttt{sailing on}, \textit{boat}) \\ \hline
(\textit{building}, \texttt{has}, \textit{building}) $\rightarrow$ (\textit{building}, \texttt{seen outside}, \textit{building}) \\ \hline
(\textit{cloud}, \texttt{in}, \textit{building}) $\rightarrow$ (\textit{cloud}, \texttt{floating through}, \textit{building}) \\ \hline
(\textit{hand}, \texttt{of}, \textit{arm}) $\rightarrow$ (\textit{hand}, \texttt{hand of}, \textit{arm}) \\ \hline
(\textit{boat}, \texttt{on}, \textit{boat}) $\rightarrow$ (\textit{boat}, \texttt{sailing on}, \textit{boat}) \\ \hline
(\textit{cloud}, \texttt{in}, \textit{sky}) $\rightarrow$ (\textit{cloud}, \texttt{floating through}, \textit{sky}) \\ \hline
(\textit{building}, \texttt{in}, \textit{background}) $\rightarrow$ (\textit{building}, \texttt{visible in}, \textit{background}) \\ \hline
(\textit{man}, \texttt{has}, \textit{arm}) $\rightarrow$ (\textit{man}, \texttt{sheltering}, \textit{arm}) \\ \hline
(\textit{man}, \texttt{wearing}, \textit{arm}) $\rightarrow$ (\textit{man}, \texttt{dressed in}, \textit{arm}) \\ \hline
(\textit{head}, \texttt{of}, \textit{arm}) $\rightarrow$ (\textit{head}, \texttt{turning}, \textit{arm}) \\ \hline
(\textit{man}, \texttt{has}, \textit{arm}) $\rightarrow$ (\textit{man}, \texttt{losing}, \textit{arm}) \\ \hline
(\textit{cloud}, \texttt{in}, \textit{airplane}) $\rightarrow$ (\textit{cloud}, \texttt{floating through}, \textit{airplane}) \\ \hline
(\textit{cloud}, \texttt{in}, \textit{background}) $\rightarrow$ (\textit{cloud}, \texttt{floating through}, \textit{background}) \\ \hline
(\textit{window}, \texttt{on}, \textit{awning}) $\rightarrow$ (\textit{window}, \texttt{on exterior of}, \textit{awning}) \\ \hline
(\textit{man}, \texttt{has}, \textit{arm}) $\rightarrow$ (\textit{man}, \texttt{pointing with}, \textit{arm}) \\ \hline
(\textit{man}, \texttt{has}, \textit{arm}) $\rightarrow$ (\textit{man}, \texttt{spreading}, \textit{arm}) \\ \hline
(\textit{window}, \texttt{on}, \textit{bus}) $\rightarrow$ (\textit{window}, \texttt{lining side of}, \textit{bus}) \\ \hline
(\textit{window}, \texttt{on}, \textit{balcony}) $\rightarrow$ (\textit{window}, \texttt{on exterior of}, \textit{balcony}) \\ \hline
(\textit{cloud}, \texttt{in}, \textit{beach}) $\rightarrow$ (\textit{cloud}, \texttt{floating through}, \textit{beach}) \\ \hline
(\textit{car}, \texttt{on}, \textit{building}) $\rightarrow$ (\textit{car}, \texttt{driving alongside}, \textit{building}) \\ \hline
(\textit{window}, \texttt{on}, \textit{building}) $\rightarrow$ (\textit{window}, \texttt{lining side of}, \textit{building}) \\ \hline
(\textit{man}, \texttt{wearing}, \textit{arm}) $\rightarrow$ (\textit{man}, \texttt{lifting up}, \textit{arm}) \\ \hline
(\textit{shirt}, \texttt{on}, \textit{arm}) $\rightarrow$ (\textit{shirt}, \texttt{worn by}, \textit{arm}) \\ \hline
(\textit{man}, \texttt{wearing}, \textit{bag}) $\rightarrow$ (\textit{man}, \texttt{wearing striped}, \textit{bag}) \\ \hline
(\textit{bottle}, \texttt{on}, \textit{bottle}) $\rightarrow$ (\textit{bottle}, \texttt{kept on}, \textit{bottle}) \\ \hline
(\textit{man}, \texttt{wearing}, \textit{background}) $\rightarrow$ (\textit{man}, \texttt{wearing striped}, \textit{background}) \\ \hline
(\textit{boy}, \texttt{wearing}, \textit{boy}) $\rightarrow$ (\textit{boy}, \texttt{striped}, \textit{boy}) \\ \hline
(\textit{cloud}, \texttt{in}, \textit{air}) $\rightarrow$ (\textit{cloud}, \texttt{floating through}, \textit{air}) \\ \hline
(\textit{woman}, \texttt{has}, \textit{arm}) $\rightarrow$ (\textit{woman}, \texttt{raising}, \textit{arm}) \\ \hline
(\textit{car}, \texttt{on}, \textit{building}) $\rightarrow$ (\textit{car}, \texttt{moving down}, \textit{building}) \\ \hline
(\textit{airplane}, \texttt{in}, \textit{airplane}) $\rightarrow$ (\textit{airplane}, \texttt{flying under}, \textit{airplane}) \\ \hline
(\textit{tile}, \texttt{on}, \textit{bathroom}) $\rightarrow$ (\textit{tile}, \texttt{fixed to}, \textit{bathroom}) \\ \hline
(\textit{arm}, \texttt{on}, \textit{arm}) $\rightarrow$ (\textit{arm}, \texttt{stretched out on}, \textit{arm}) \\ \hline
(\textit{cloud}, \texttt{in}, \textit{arm}) $\rightarrow$ (\textit{cloud}, \texttt{floating through}, \textit{arm}) \\ \hline
(\textit{head}, \texttt{of}, \textit{arm}) $\rightarrow$ (\textit{head}, \texttt{belonging to}, \textit{arm}) \\ \hline
(\textit{man}, \texttt{wearing}, \textit{arm}) $\rightarrow$ (\textit{man}, \texttt{kicking up}, \textit{arm}) \\ \hline
(\textit{airplane}, \texttt{on}, \textit{airplane}) $\rightarrow$ (\textit{airplane}, \texttt{taking off from}, \textit{airplane}) \\ \hline
(\textit{sign}, \texttt{on}, \textit{building}) $\rightarrow$ (\textit{sign}, \texttt{strapped}, \textit{building}) \\ \hline
(\textit{man}, \texttt{wearing}, \textit{air}) $\rightarrow$ (\textit{man}, \texttt{wearing striped}, \textit{air}) \\ \hline
(\textit{sign}, \texttt{on}, \textit{arrow}) $\rightarrow$ (\textit{sign}, \texttt{strapped}, \textit{arrow}) \\ \hline
(\textit{window}, \texttt{on}, \textit{building}) $\rightarrow$ (\textit{window}, \texttt{adorning}, \textit{building}) \\ \hline
(\textit{cat}, \texttt{has}, \textit{cat}) $\rightarrow$ (\textit{cat}, \texttt{possesses}, \textit{cat}) \\ \hline
(\textit{cloud}, \texttt{in}, \textit{boat}) $\rightarrow$ (\textit{cloud}, \texttt{floating through}, \textit{boat}) \\ \hline
(\textit{person}, \texttt{has}, \textit{arm}) $\rightarrow$ (\textit{person}, \texttt{stretching out}, \textit{arm}) \\ \hline
(\textit{clock}, \texttt{on}, \textit{building}) $\rightarrow$ (\textit{clock}, \texttt{attached to side of}, \textit{building}) \\ \hline
(\textit{train}, \texttt{on}, \textit{building}) $\rightarrow$ (\textit{train}, \texttt{switching}, \textit{building}) \\ \hline
(\textit{woman}, \texttt{has}, \textit{arm}) $\rightarrow$ (\textit{woman}, \texttt{combing}, \textit{arm}) \\ \hline
(\textit{boy}, \texttt{wearing}, \textit{arm}) $\rightarrow$ (\textit{boy}, \texttt{striped}, \textit{arm}) \\ \hline
(\textit{window}, \texttt{on}, \textit{building}) $\rightarrow$ (\textit{window}, \texttt{on the side of}, \textit{building}) \\ \hline
(\textit{window}, \texttt{of}, \textit{building}) $\rightarrow$ (\textit{window}, \texttt{on exterior of}, \textit{building}) \\ \hline
(\textit{arrow}, \texttt{on}, \textit{arrow}) $\rightarrow$ (\textit{arrow}, \texttt{printed}, \textit{arrow}) \\ \hline
(\textit{woman}, \texttt{wearing}, \textit{arm}) $\rightarrow$ (\textit{woman}, \texttt{wearing striped}, \textit{arm}) \\ \hline
(\textit{head}, \texttt{of}, \textit{arm}) $\rightarrow$ (\textit{head}, \texttt{turned to}, \textit{arm}) \\ \hline
(\textit{branch}, \texttt{on}, \textit{branch}) $\rightarrow$ (\textit{branch}, \texttt{sticking up on}, \textit{branch}) \\ \hline
(\textit{man}, \texttt{on}, \textit{arm}) $\rightarrow$ (\textit{man}, \texttt{swimming with}, \textit{arm}) \\ \hline
(\textit{wing}, \texttt{on}, \textit{airplane}) $\rightarrow$ (\textit{wing}, \texttt{on left side of}, \textit{airplane}) \\ \hline
(\textit{mountain}, \texttt{in}, \textit{background}) $\rightarrow$ (\textit{mountain}, \texttt{visible in}, \textit{background}) \\ \hline
(\textit{bowl}, \texttt{on}, \textit{bowl}) $\rightarrow$ (\textit{bowl}, \texttt{placed on}, \textit{bowl}) \\ \hline
(\textit{cloud}, \texttt{in}, \textit{blue sky}) $\rightarrow$ (\textit{cloud}, \texttt{floating through}, \textit{blue sky}) \\ \hline
(\textit{window}, \texttt{on}, \textit{arrow}) $\rightarrow$ (\textit{window}, \texttt{on exterior of}, \textit{arrow}) \\ \hline
(\textit{foot}, \texttt{of}, \textit{arm}) $\rightarrow$ (\textit{foot}, \texttt{belonging to}, \textit{arm}) \\ \hline
(\textit{toilet}, \texttt{in}, \textit{bathroom}) $\rightarrow$ (\textit{toilet}, \texttt{installed in}, \textit{bathroom}) \\ \hline
(\textit{sign}, \texttt{on}, \textit{building}) $\rightarrow$ (\textit{sign}, \texttt{anchored to}, \textit{building}) \\ \hline
(\textit{wall}, \texttt{on}, \textit{building}) $\rightarrow$ (\textit{wall}, \texttt{making up}, \textit{building}) \\ \hline
(\textit{kite}, \texttt{in}, \textit{air}) $\rightarrow$ (\textit{kite}, \texttt{flying through}, \textit{air}) \\ \hline
(\textit{leaf}, \texttt{on}, \textit{building}) $\rightarrow$ (\textit{leaf}, \texttt{growing on}, \textit{building}) \\ \hline
(\textit{man}, \texttt{wearing}, \textit{arm}) $\rightarrow$ (\textit{man}, \texttt{adjusting}, \textit{arm}) \\ \hline
(\textit{person}, \texttt{wearing}, \textit{arm}) $\rightarrow$ (\textit{person}, \texttt{striped}, \textit{arm}) \\ \hline
(\textit{tile}, \texttt{on}, \textit{bathroom}) $\rightarrow$ (\textit{tile}, \texttt{installed on}, \textit{bathroom}) \\ \hline
(\textit{bag}, \texttt{on}, \textit{bag}) $\rightarrow$ (\textit{bag}, \texttt{kept in}, \textit{bag}) \\ \hline
(\textit{sky}, \texttt{in}, \textit{sky}) $\rightarrow$ (\textit{sky}, \texttt{stretched across}, \textit{sky}) \\ \hline
(\textit{bear}, \texttt{has}, \textit{bear}) $\rightarrow$ (\textit{bear}, \texttt{scratching}, \textit{bear}) \\ \hline
(\textit{line}, \texttt{on}, \textit{building}) $\rightarrow$ (\textit{line}, \texttt{painted in}, \textit{building}) \\ \hline
(\textit{man}, \texttt{wearing}, \textit{building}) $\rightarrow$ (\textit{man}, \texttt{wearing striped}, \textit{building}) \\ \hline
(\textit{book}, \texttt{on}, \textit{book}) $\rightarrow$ (\textit{book}, \texttt{arranged on}, \textit{book}) \\ \hline
(\textit{wall}, \texttt{near}, \textit{building}) $\rightarrow$ (\textit{wall}, \texttt{making up}, \textit{building}) \\ \hline
(\textit{window}, \texttt{on}, \textit{advertisement}) $\rightarrow$ (\textit{window}, \texttt{lining side of}, \textit{advertisement}) \\ \hline
(\textit{sink}, \texttt{in}, \textit{bathroom}) $\rightarrow$ (\textit{sink}, \texttt{mounted in}, \textit{bathroom}) \\ \hline
(\textit{wave}, \texttt{in}, \textit{arm}) $\rightarrow$ (\textit{wave}, \texttt{cresting in}, \textit{arm}) \\ \hline
(\textit{window}, \texttt{in}, \textit{building}) $\rightarrow$ (\textit{window}, \texttt{on exterior of}, \textit{building}) \\ \hline
(\textit{blanket}, \texttt{on}, \textit{bed}) $\rightarrow$ (\textit{blanket}, \texttt{laying over}, \textit{bed}) \\ \hline
(\textit{man}, \texttt{wearing}, \textit{backpack}) $\rightarrow$ (\textit{man}, \texttt{carrying}, \textit{backpack}) \\ \hline
(\textit{hair}, \texttt{of}, \textit{arm}) $\rightarrow$ (\textit{hair}, \texttt{on head of}, \textit{arm}) \\ \hline
(\textit{kite}, \texttt{in}, \textit{beach}) $\rightarrow$ (\textit{kite}, \texttt{kite in}, \textit{beach}) \\ \hline
(\textit{man}, \texttt{wearing}, \textit{arm}) $\rightarrow$ (\textit{man}, \texttt{dressed}, \textit{arm}) \\ \hline
(\textit{boy}, \texttt{has}, \textit{arm}) $\rightarrow$ (\textit{boy}, \texttt{outstretched}, \textit{arm}) \\ \hline
(\textit{picture}, \texttt{on}, \textit{bed}) $\rightarrow$ (\textit{picture}, \texttt{framed on}, \textit{bed}) \\ \hline
(\textit{window}, \texttt{on}, \textit{banner}) $\rightarrow$ (\textit{window}, \texttt{on exterior of}, \textit{banner}) \\ \hline
(\textit{arm}, \texttt{of}, \textit{arm}) $\rightarrow$ (\textit{arm}, \texttt{belonging to}, \textit{arm}) \\ \hline
(\textit{man}, \texttt{on}, \textit{board}) $\rightarrow$ (\textit{man}, \texttt{going off}, \textit{board}) \\ \hline
(\textit{shadow}, \texttt{on}, \textit{arm}) $\rightarrow$ (\textit{shadow}, \texttt{cast over}, \textit{arm}) \\ \hline
(\textit{arm}, \texttt{of}, \textit{arm}) $\rightarrow$ (\textit{arm}, \texttt{around neck of}, \textit{arm}) \\ \hline
\end{longtable}
}




\end{document}